\documentclass{article}

\usepackage[preprint,nonatbib]{neurips_2025}
\usepackage{graphicx}
\usepackage{subcaption}
\usepackage{array}
\usepackage{caption}
\usepackage{enumitem}
\usepackage{subcaption}
\usepackage{makecell} 
\usepackage{pgfplots}
\pgfplotsset{compat=1.18}
\usepackage{tikz}
\usetikzlibrary{patterns, decorations.text}
\usepackage{algorithm}
\usepackage{algpseudocode}
\usepackage{multirow}  
\usepackage{xcolor}
\usepackage[utf8]{inputenc} 
\usepackage[T1]{fontenc}    
\usepackage{hyperref}       
\usepackage{url}            
\usepackage{booktabs}       
\usepackage{amsfonts}       
\usepackage{nicefrac}       
\usepackage{microtype}      
\usepackage{xcolor}         
\usepackage{tikz}
\usepackage{pgfplots}
\usepackage{enumitem}
\usepackage{amsmath}
\usepackage{booktabs}
\usepackage{siunitx} 
\usepackage[export]{adjustbox}
\title{
\includegraphics[height=3.5em, width=3.5em]{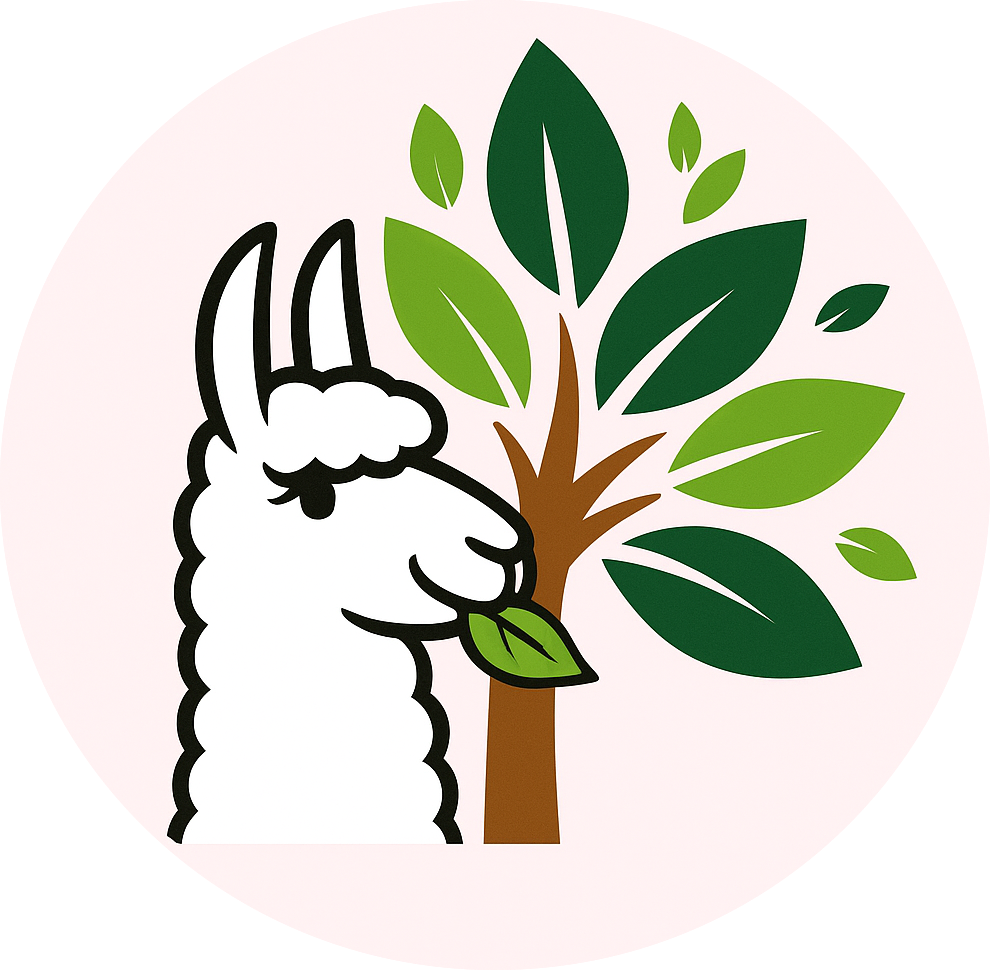} \\ 
SpecMemo: Speculative Decoding is in Your Pocket}

\author{%
  Selin Yildirim\thanks{First author.} \hspace{2em} Deming Chen \\
  University of Illinois Urbana-Champaign \\
  \texttt{\{seliny2,dchen\}@illinois.edu}
}

\begin{document}

\maketitle

\begin{abstract}
 Recent advancements in speculative decoding have demonstrated considerable speedup across a wide array of large language model (LLM) tasks. Speculative decoding inherently relies on sacrificing extra memory allocations to generate several candidate tokens, of which acceptance rate drives the speedup. However, deploying speculative decoding on memory-constrained devices, such as mobile GPUs, remains as a significant challenge in real-world scenarios. In this work, we present a device-aware inference engine named SpecMemo that can smartly control memory allocations at finer levels to enable multi-turn chatbots with speculative decoding on such limited memory devices. Our methodology stems from theoretically modeling memory footprint of speculative decoding to determine a lower bound on the required memory budget while retaining speedup. SpecMemo empirically acquires a careful balance between minimizing redundant memory allocations for rejected candidate tokens and maintaining competitive performance gains from speculation. Notably, with SpecMemo's memory management, we maintain 96\% of overall throughput from speculative decoding on MT-Bench\cite{mtbench}, with reduced generation-memory by 65\% on single Nvidia Titan RTX \cite{nvidia_titan}. Given multiple constrained GPUs, we build on top of previous speculative decoding architectures to facilitate big-model inference by distributing Llama-2-70B-Chat\cite{llama70b,llama2} model, on which we provide novel batched speculative decoding to increase usability of multiple small server GPUs. This novel framework demonstrates 2x speedup over distributed and batched vanilla decoding with the base model on eight AMD MI250 GPUs\cite{amd_mi250}. Moreover, inference throughput increases remarkably 8x with batch size 10. Our work contributes to democratized LLM applications in resource-constrained environments, providing a pathway for faster and cheaper deployment of real-world LLM applications with robust performance.
\end{abstract}

\section{Introduction}
\begin{figure}[h!]
    \centering
\includegraphics[width=\linewidth]{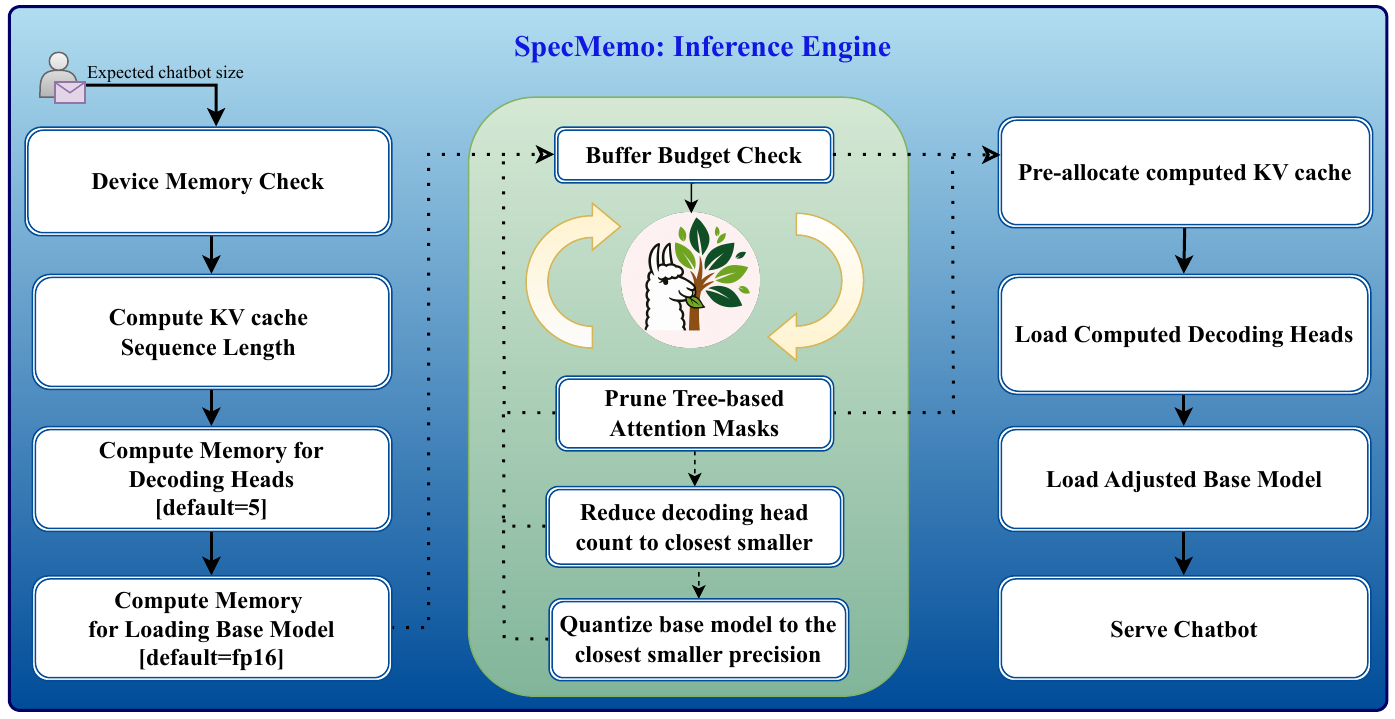}
    \caption{SpecMemo Inference Engine Architecture}
    \label{fig:specmemo}
\end{figure}
The rapid advancements in large language models (LLMs) \cite{vaswani2023attentionneed} have opened up new possibilities for a variety of natural language processing tasks, from text generation to complex decision-making. 
Speculative decoding\cite{chen2023acceleratinglargelanguagemodel,cai2024medusasimplellminference,specinfer,li2025eaglespeculativesamplingrequires,wu2025specrouteradaptiveroutingmultilevel,wang2024continuousspeculativedecodingautoregressive,jang2025lantern,sun2025blockverificationacceleratesspeculative,yang2025longspeclongcontextspeculativedecoding,bachmann2025judgedecodingfasterspeculative} has emerged as an effective technique to address autoregressive nature of text generation, enabling faster decoding by leveraging redundant computations and extra memory allocations. Despite its speedup potential, existing speculative decoding methods are not the best fits for memory constrained environments due to its extra memory usage, thus mostly targeting data-center GPUs. Deployment of such solution on consumer-grade hardware, such as GPUs integrated into laptops and edge devices, remains a significant challenge when not addressed carefully, due to the substantial memory demand at inference-time. This aspect make existing speculative decoding methods impractical for the use in real-world applications on consumer devices.

To make fast and accurate inference cheap and accessible to all, we study the limitations of serving speculative decoding on memory-constrained GPUs. As example speculative decoding application, we built on top of Medusa\cite{cai2024medusasimplellminference} due to its following advantages. Medusa\cite{cai2024medusasimplellminference} eliminates the need for a draft model, which simplifies deployment, reduces memory footprint and removes the need for cross-model verification with higher acceptance rates between base model and parallel decoding heads, which are fine-tuned on the base model. We first provide a theoretical analysis on memory footprint of Medusa-based\cite{cai2024medusasimplellminference} speculative decoding applications in Section \ref{chapter:3.1}, which by design uses a central base model and parallel speculative decoding heads. We study and model the run-time memory allocations performed by speculative decoding, which is essential for reducing out-of-memory (OOM) errors. We propose controlling the buffer allocations for token sampling phase ahead of time to prevent errors, however, such an approach is challenging in retaining acceptance frequency of candidate tokens. While mitigating memory demands of sampling, it is also crucial to retain speedup gains from sampling phase. 

We demonstrate that second challenge in memory budgeting of speculative decoding is pre-allocating a precise space for the KV cache. Larger spaces for KV cache are typically expected in long context query processing, such as real-world multi-turn chatbots with an ability to remember previous user prompts. The more queries to be served in a multi-turn chatbot, the larger KV cache allocation would be needed, which stresses the constrained GPUs more. Existing works set the pre-allocated sequence length of KV cache to the model's maximum embedding space, due to expected model failures after exceeding trained context window. However, it is often resource-wasting and latency-inducing to move larger chunks of unused KV cache data around GPUs during speculation. Therefore, it is important to closely approximate useful sequence length and pre-allocate just enough memory for the KV cache to serve chatbots without failure. 

\begin{figure*}[h!]
\centering
    \begin{tabular}{cccc}
         \subcaptionbox{Speculation Step: 1}
        {\includegraphics[width=0.44\linewidth, height=0.16\textheight]{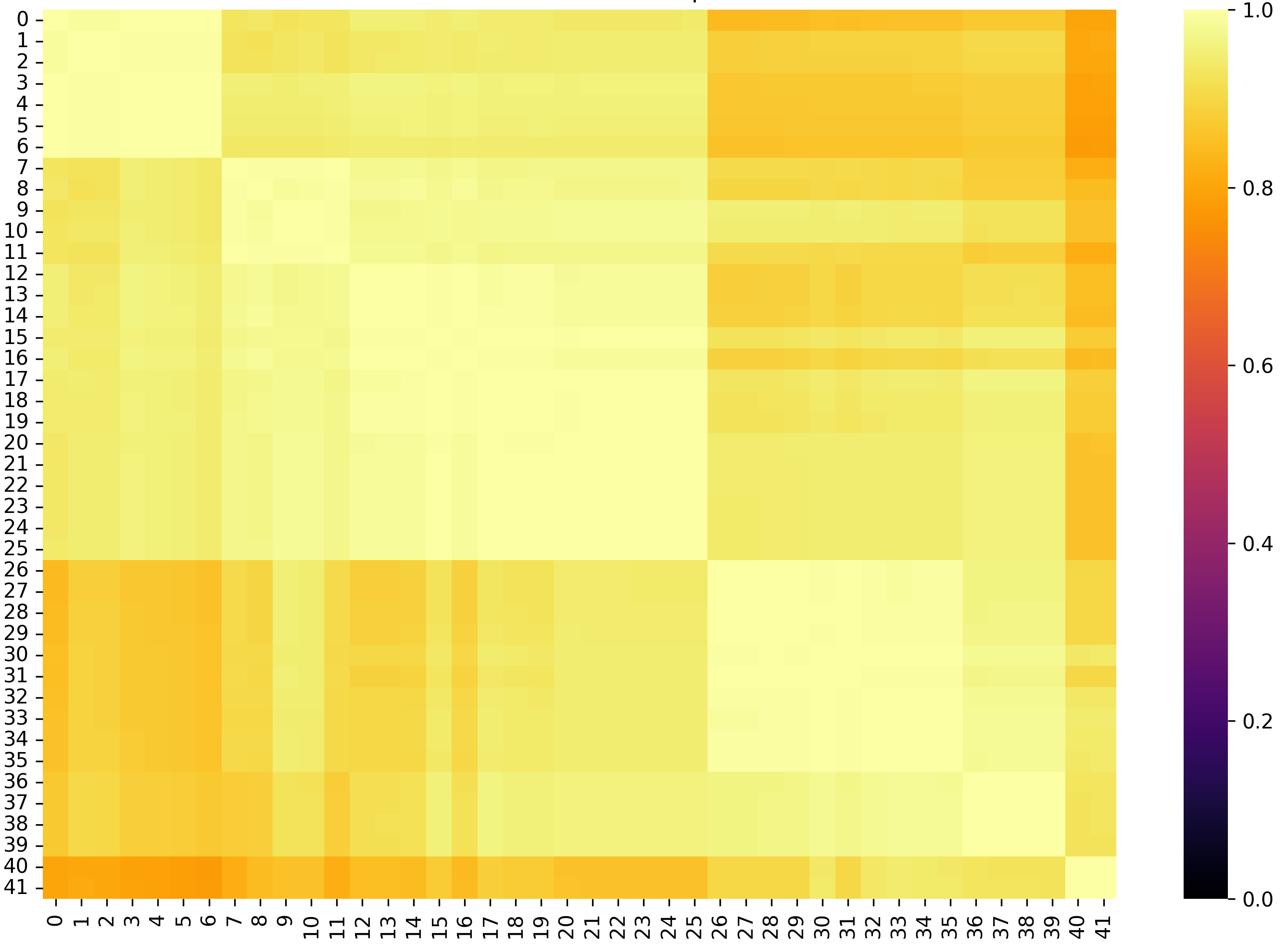}} &
        \subcaptionbox{Speculation Step: 150}
       {\includegraphics[width=0.44\linewidth, height=0.16\textheight]{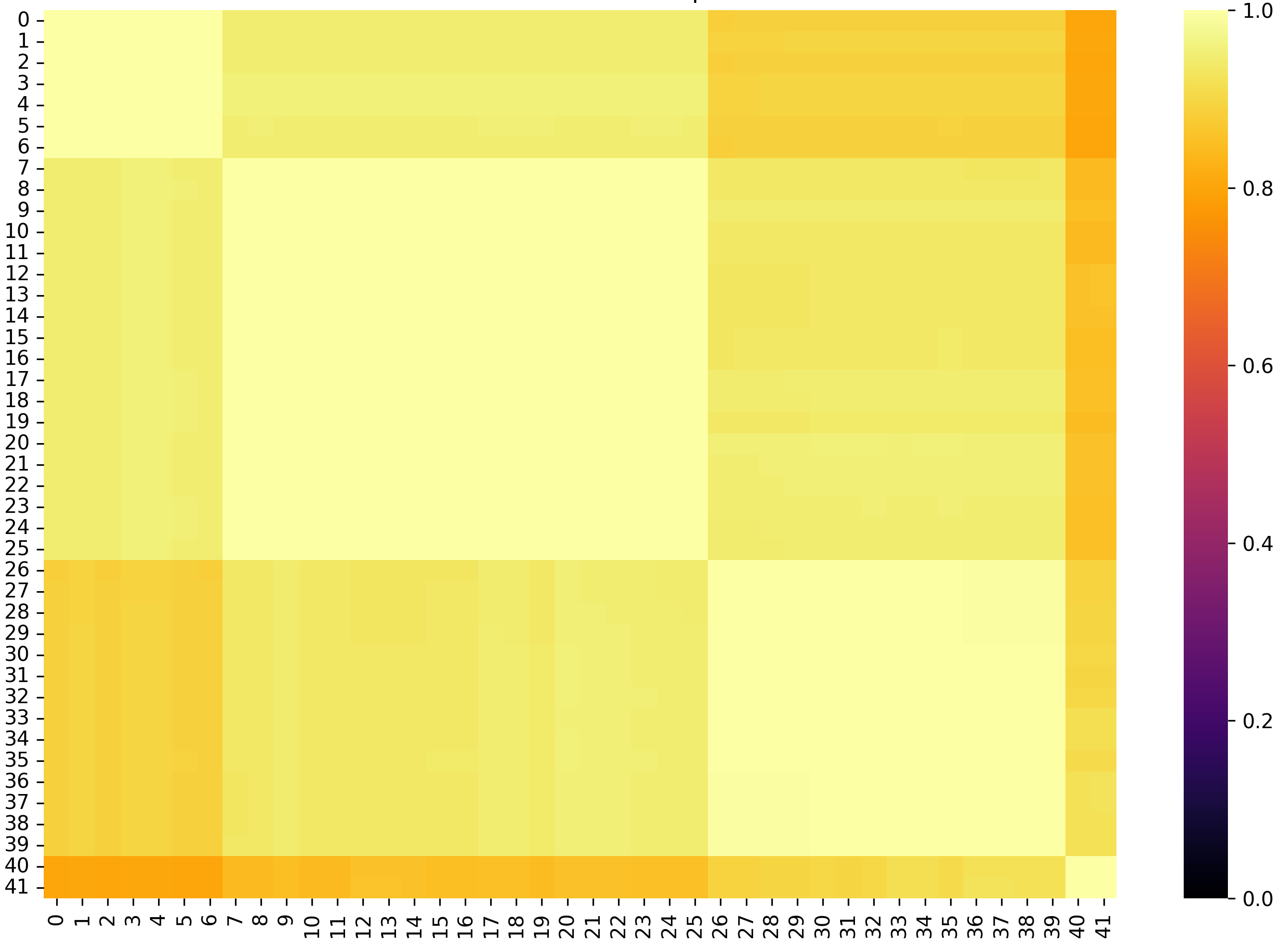}} \\
    \end{tabular}
    \caption{Pre-verification cosine similarity of 42 candidate sequences present in original Medusa\cite{cai2024medusasimplellminference} mask during story generation task with Vicuna-7B \cite{vicuna7b}. Further details are shared in Appendix Section \ref{chapter:appendtree}}
    \label{fig:similarty}
\end{figure*}

To address these challenges on a constrained GPU, we propose a training-free novel inference engine named SpecMemo that can control memory allocations of speculative decoding at finer levels before generation, to maintain high generation accuracy and speedup over vanilla decoding. The key idea behind our budgeting approach exploits the observation that candidate sequences significantly share higher cosine similarities before verification as shown in Figure \ref{fig:similarty}. Our observation unlocks the opportunity to eliminate some portion of sequences via tree pruning, which dramatically reduces the overall buffer allocations for sampling throughout the generation. Therefore, SpecMemo utilizes tree-based attention\cite{vaswani2023attentionneed} mask customization via heuristic-based pruning strategies to reduce generation-time memory footprint and retain high speedup. SpecMemo operates on a plug-and-play algorithm that can be easily integrated into several state-of-the-art speculative decoding methods to balance performance trade-offs, making it a compelling solution for memory-constrained environments. Details on SpecMemo's memory-friendly optimizations are presented in Section \ref{chapter-tree}.

To the best of our knowledge, SpecMemo is the first to propose and implement distributed and batched speculative decoding for scenarios involving multiple constrained GPUs. SpecMemo unlocks big-model inference functionality (where model does not fit on a single GPU) by equally splitting layers of the base model across multiple memory-limited GPUs. To mitigate inter-GPU communication latencies from distribution, we also develop a novel batched speculative decoding strategy on top of the distributed base model. Speculative decoding poses unique batching challenges due to the variable-length acceptance of candidate tokens at each generation step, unlike standard autoregressive decoding. Additionally, performance is non-trivial to maintain with batched speculative decoding, as it incurs extra memory overhead from padding position IDs and KV caches compared to vanilla batching. Despite these challenges, our approach achieves an 8x throughput improvement (up to batch size 10) and 2x speedup over distributed and batched autoregressive decoding.

Combining the two novelties, our work enables LLMs to be deployed more effectively on consumer GPUs, such as those found in Windows\cite{windows} laptops with Nvidia GeForce GPUs \cite{nvidia_rtx}, which typically have limited memory resources. By striking an optimal balance between memory optimization, accuracy and decoding speed, our method paves the way for the deployment of large language models in real-time, resource-constrained environments, without compromising on performance. This paper contributes to the ongoing effort to bring powerful LLMs to the edge, making them more accessible and efficient for a wide range of practical use cases.

Contributions of this paper are summarized as follows.
\begin{itemize}[noitemsep, topsep=0pt, leftmargin=*, labelsep=0.5em]
\item \textbf{Tree-based mask pruning for speculative decoding:} We analyze the structure of optimal tree-based attention masks that improve acceptance rates—central to achieving speedup in speculative decoding. Building on this insight, we develop a device-aware inference engine, SpecMemo, which explores variable-sized tree-based masks to be used in the sampling phase. This approach enables up to 65\% reduction in runtime memory usage on resource-constrained single-GPU setups, while maintaining 96\% of the speedup.
\item \textbf{Automatic inference hyperparameter adjustments:} SpecMemo includes a predictive mechanism for automatically selecting optimal hyperparameters before generation, such as the number of parallel decoding heads and KV cache allocation. Combined with adaptive mask sizing, this allows SpecMemo to automatically guarantee serving a target number of user queries (e.g., in chatbot scenarios) without exceeding memory limits (OOM).
\item \textbf{Scalable speculative decoding on constrained multi-GPU systems:} SpecMemo extends speculative decoding to large-scale models that do not fit on a single GPU by distributing the verifier model across multiple memory-constrained GPUs. Additionally, we incorporate a novel batching strategy that boosts inference throughput by 8× with a batch size of 10, and accelerates up to 2× over distributed, batched base model.
\item \textbf{Evaluation of the speculative decoding design space:} We conduct an extensive performance evaluation over a three-dimensional space that includes varying tree-based mask structures and abovementioned inference hyperparameters. This analysis reveals key “sweet spots” for maximizing speculative decoding efficiency under various deployment scenarios and guides our memory optimizations.
\end{itemize}

\section{Related works}
Tree-based speculative decoding methods\cite{cai2024medusasimplellminference, specinfer}offer improved acceptance rates ($\alpha$) compared to earlier methods, speculative sampling \cite{chen2023acceleratinglargelanguagemodel} and chain-based verification \cite{wu2025specrouteradaptiveroutingmultilevel}. While tree-based speculative decoding offer longer sequences in a single decoding pass, it often causes greater overall computational and memory overhead due to rejected tokens on the tree. Notable implementations that leverage tree-based attention masks include Medusa\cite{cai2024medusasimplellminference} and Eagle\cite{li2025eaglespeculativesamplingrequires}, both of which accelerate generation but do not address memory budget constraints on limited-resource GPUs, particularly causing frequent out-of-memory (OOM) failures under various inference configurations at runtime. Eagle-2\cite{li2024eagle2fasterinferencelanguage} constructs its attention tree dynamically using token-probability-based filtering at each generation step. Although this improves generation speed over Medusa \cite{cai2024medusasimplellminference}, the dynamic token probability exploration still introduces significant memory demands, especially in chatbot applications. 

Both Medusa \cite{cai2024medusasimplellminference} and Eagle \cite{li2025eaglespeculativesamplingrequires} rely on pre-allocated KV caches that match the model’s full context length—an inefficient design for memory-constrained environments. Recent KV cache compression techniques \cite{zhang2023h2oheavyhitteroracleefficient,wang2024modeltellsmergeadaptive,ge2024modeltellsdiscardadaptive,han2024lminfinitezeroshotextremelength,cai2025pyramidkvdynamickvcache,juravsky2024hydragenhighthroughputllminference} for vanilla decoding are non-trivial to extend to speculative decoding: Each speculative decoding head must retain high numerical precision to pass cumulative verification. This kind of optimization typically requires additional precision adjustments through fine tuning of speculative heads. 

The primary challenge in speculative decoding remains designing a precise attention mask that includes only to-be-verified tokens. Naive (full) tree construction is highly infeasible due to the massive memory and computational costs. Therefore, Medusa \cite{cai2024medusasimplellminference} ships with a small set of static, pre-built and pruned attention trees targeting specific models like Vicuna\cite{vicuna7b} and Zephyr\cite{zephyr}. Remaining challenge in static Medusa \cite{cai2024medusasimplellminference} masks are their likelihood to create OOM failures at run-time on memory-limited GPUs due to their relatively large node counts. In contrast, SpecMemo explores custom attention masks that can be seamlessly integrated into the sampling phases of both Medusa and Eagle\cite{li2025eaglespeculativesamplingrequires} on various GPU sizes. SpecMemo inference engine offers a set of effective memory optimizations for Medusa-style\cite{cai2024medusasimplellminference,specinfer} speculative decoding systems, which rely on a centralized base model and multiple parallel speculative decoding heads. Similar work \cite{sequoia} on optimizing attention-tree masks relies on model offloading technique on Nvidia L40 data-center GPU \cite{l40} that incurs inter-device communication latencies. Paper \cite{yin2024theoreticalperspectivespeculativedecoding} analyzes optimality of speculative decoding theoretically without investigating its memory footprint. 

Existing batching approach for speculative decoding is provided in \cite{qian2024bassbatchedattentionoptimizedspeculative}. However, this work also targets draft-model-based architecture and evaluates latency on Nvidia A100 GPU \cite{nvidia_a100} with large memory resources. They lack big-model inference support while evaluating their approach with smaller LLMs. On the other hand, \cite{su2023synergyspeculativedecodingbatching} offers searching an adaptive speculation length for a batch, however this approach is highly inefficient under large batches if the user batch size is constant (uniform traffic). Therefore, this approach also fails offering a solution for consumer-grade GPU inference where user traffic is uniform and low. Moreover, our batched speculative decoding batching optionally facilitates big model inference across many GPUs, wherease \cite{su2023synergyspeculativedecodingbatching} focuses on small draft models.

\section{SpecMemo}
\subsection{Formulating memory requirements} \label{chapter:3.1}
The memory consumption of speculative decoding can be categorized into two main components, which we empirically identify as the primary contributors to both static and dynamic memory allocations. 
\paragraph{KV cache allocation:} Speculative decoding pre-allocates a KV cache to use it as a scratch space and frequently update during verification stage. Therefore, it is more efficient to compute a precise sequence length that helps serve certain number of user queries and minimize memory waste. KV cache allocation for serving a chatbot, with $n$ user queries of each $m$=max\_token\_length, is modeled as follows by also involving the model-dependent parameters and precision.
\begin{equation}
\text{Memory}_{KV_{nm}} = 2*h*b*k*x*p
\end{equation}, where $h$ is the number of hidden layers, $b$ is batch size, $k$ is the number of KV heads, $x=m*n$ is computed sequence length, $d$ is hidden dimension size per attention head, $p=$ is model precision. 

\paragraph{Buffer allocations:} \label{chapter:3.1.2} Speculative decoding incurs run-time memory allocations for managing internal buffers (i.e, sampling logits from heads, verifying sequences with tree mask, accepting a sequence based on an entropy threshold), which affects the feasibility of serving queries on a limited environment. Assuming $L$-many parallel decoding heads (equals to the number of tree levels on the attention mask, $l$), a tree arity-$k$ (equals to top-$k$ token sampling from heads), number of nodes $N$ and total candidate sequences $S$, tree-based attention mask size is formulated as follows. Prior to verification, attention mask assumes full acceptance for each sequence.
\begin{equation}
\text{$N$} = \sum_{i=0}^{l} k^i, l \in \{1, 2, \dots, 5\}, k \in \{5,10\},       
    \text{Attention Mask$(N,S,l)$} = \bigcup_{i=0}^{S} \bigcup_{j=0}^{l} \text{Node}_{i,j}=1
\label{eqn:tree}
\end{equation} 

Assuming a tree-based attention mask as formulated in Equation \ref{eqn:tree}, intermediate buffer allocations at speculative decoding runtime are formulated in Equation \ref{eqn:three}: The first term covers the space allocated by sampled logits, which are as many as attention node count of the tree mask. Buffers modeled in the second and third term are used to retrieve sequences from the tree mask and compute the acceptance length ($\tau$) of each sequence, respectively.
\begin{equation} \label{eqn:three}
\text{Memory}_{buffers} = b * N * w  + b* S * l * w + b* S * l * l * w \text{,where $w$ is vocabulary size}
\end{equation} 

\begin{figure}[h!]
    \centering
    \begin{minipage}[t]{0.49\linewidth}
        \includegraphics[height=5cm, width=\linewidth]{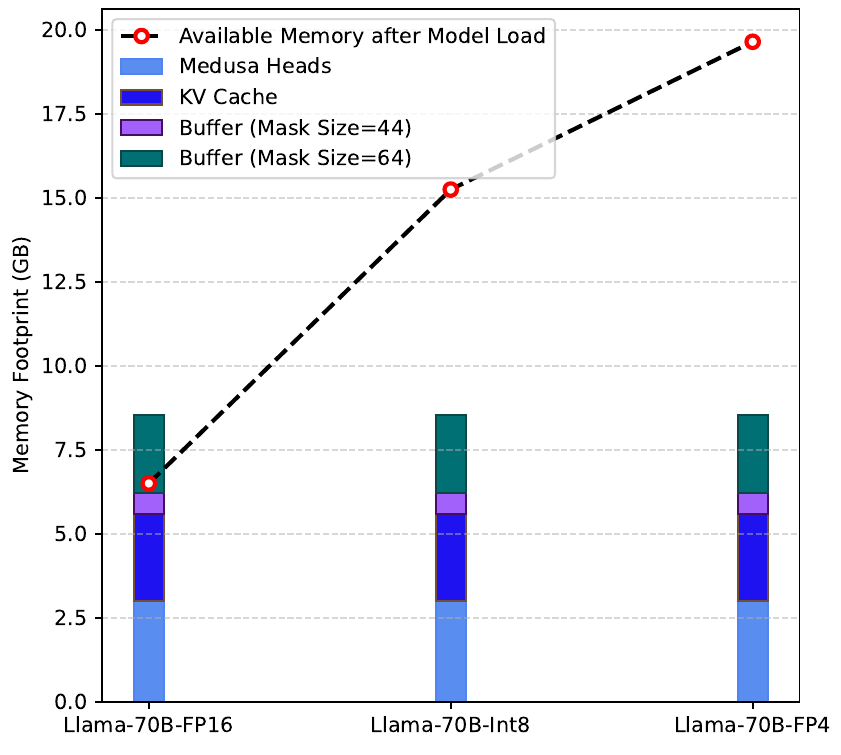}
    \caption{Memory breakdown by the theoretical guide provided in Equation \ref{totalmem-eqn}. Attention mask choice of size $64$ causes OOM in FP16, whereas the smaller mask can succeed.}
    \label{fig:failuredemo}
    \end{minipage}
    \hfill
\begin{minipage}[t]{0.49\linewidth}
\hspace{-0.3cm}
\raisebox{0.2cm}{
\begin{tikzpicture}[scale=0.45]
\begin{axis}[
    xlabel = {\# of Tokens},
    ylabel = {Memory Allocation (GB)},
    grid = major,
    width=12cm, 
    height=10cm, 
    legend style={
        at={(0.5,-0.15)}, 
        anchor=north, 
        legend columns=5,
        column sep=1ex 
    }, 
]
\addplot+[
    mark=*, 
    mark options={fill=black}, 
    line width=1pt, 
    color = black,
] coordinates {(0, 16) (0, 17)}; 
\addlegendentry{Pre-Allocating KV Cache}
\node at (axis cs:0, 17) [anchor=north east,font=\small] {1 GB};
\addplot+[
    mark=*, 
    mark options={fill=cyan}, 
    color=blue!70!black, 
    line width=1pt, 
] coordinates {(0, 17) (1226, 18.3) }; \addlegendentry{1 Branch}
\addplot+[
    mark=*, 
    mark options={fill=yellow!80!black}, 
    color=green!70!black, 
    line width=1pt, 
] coordinates {(0, 17) (1490, 18.9) }; 
\addlegendentry{Quarter Tree}
\addplot+[
    mark=*, 
    mark options={fill=pink!50}, 
    color=magenta , 
    line width=1pt, 
] coordinates {(0, 17) (1432, 18.5) }; 
\addlegendentry{Half Tree}
\addplot+[
    mark=*, 
    mark options={fill=brown}, 
    color=purple!70!black, 
    line width=1pt, 
] coordinates {(0, 17) (1304, 18.1) }; \addlegendentry{Original Tree}

\addplot+[
    mark=*, 
    mark options={fill=black}, 
    line width=1pt, 
    color = black,
] coordinates {(0, 16) (0, 21.2)}; 
\node at (axis cs:0, 21.2) [anchor=north east,font=\small] {5 GB};
\addplot+[
    mark=*, 
    mark options={fill=cyan}, 
    color=blue!70!black, 
    line width=1pt, 
] coordinates {(0, 21.2) (1624, 22.3) }; 
\addplot+[
    mark=*, 
    mark options={fill=yellow!80!black}, 
    color=green!70!black, 
    line width=1pt, 
] coordinates {(0, 21.2) (1781, 22.9) }; 
\addplot+[
    mark=*, 
    mark options={fill=pink!50}, 
    color=magenta , 
    line width=1pt, 
] coordinates {(0, 21.2) (1708, 22.7) };
\addplot+[
    mark=*, 
    mark options={fill=brown}, 
    color=purple!70!black, 
    line width=1pt, 
] coordinates {(0, 21.2) (1566, 22.7) }; 

\addplot+[
    mark=*, 
    mark options={fill=black}, 
    line width=1pt, 
    color = black,
] coordinates {(0, 16) (0, 23.5)}; 
\node at (axis cs:0, 23.5) [anchor=north east,font=\small] {7 GB};
\addplot+[
    mark=*, 
    mark options={fill=cyan}, 
    color=blue!70!black, 
    line width=1pt, 
] coordinates {(0, 23.5) (1624 , 23.8) }; 
\addplot+[
    mark=*, 
    mark options={fill=yellow!80!black}, 
    color=green!70!black, 
    line width=1pt, 
] coordinates {(0, 23.5) (1490 , 23.7) }; 
\addplot+[
    mark=*, 
    mark options={fill=pink!50}, 
    color=magenta , 
    line width=1pt, 
] coordinates {(0, 23.5) (1456, 23.7) };
\addplot+[
    mark=*, 
    mark options={fill=brown}, 
    color=purple!70!black, 
    line width=1pt, 
] coordinates {(0, 23.5) (1570 , 23.9) };

\end{axis}
\end{tikzpicture}
}
\caption{Correlation between memory use and maximum generation length of Medusa\cite{cai2024medusasimplellminference} with Vicuna-7B\cite{vicuna7b}. Original tree is the same mask with size $64$ in Figure \ref{fig:failuredemo}, which is shipped by Medusa\cite{cai2024medusasimplellminference}.}
\label{fig:genlen-vs-memory}
\end{minipage}
\end{figure}

\paragraph{Total memory:} Assuming that space reserved for the intermediate buffers is reused across generation steps, overall memory footprint of generating one query by several iterations is,
\begin{equation}
\text{Memory}_{heads} = 0.6 GB * l\text{, since size of Medusa\cite{cai2024medusasimplellminference} heads are constant}
\end{equation} 
\begin{equation}
\text{Memory}_{base\_model} = B * p \text{, where $B$ is model parameter size}
\end{equation} 
\begin{equation} \label{totalmem-eqn}
\text{Memory}_{total} = \text{Memory}_{(base\_model + heads + KV_{n*m} * p)} + \text{Memory}_{buffers} * p
\end{equation} 
Following this guide, we highlight the importance of minimizing attention tree mask’s size in specific inference scenarios as shown in Figure \ref{fig:failuredemo}, where the runtime buffers become the bottleneck of generation. Thus, SpecMemo crafts a memory-friendly tree-based attention mask based on real-time GPU budget for chosen inference hyper-parameters. Figure \ref{fig:genlen-vs-memory} presents improved longest possible generation lengths under partitioned tree mask sizes on Nvidia Titan RTX \cite{nvidia_titan}. Using medium-size attention masks (quarter and half tree) consistently results in longer token generation. 

\begin{figure}
\centering
\begin{minipage}[t]{0.39\linewidth}
        \includegraphics[width=\linewidth]{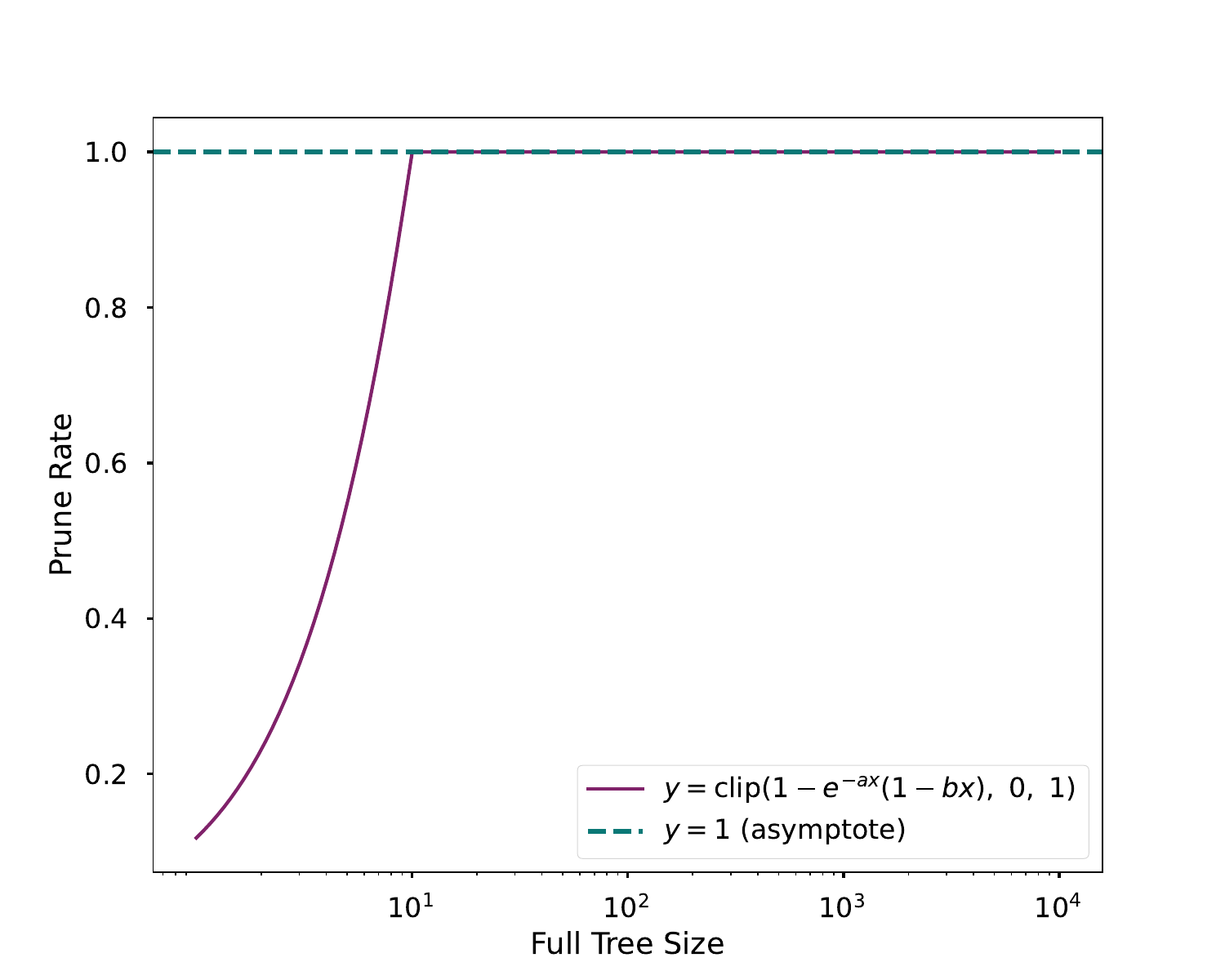}
    \caption{Tree pruning function for a growing full attention tree with arity $k=10$.}
    \label{fig:prunefunc}
\end{minipage}
\hspace{0.3cm}
\begin{minipage}[t]{0.45\linewidth}
    \includegraphics[width=\linewidth]{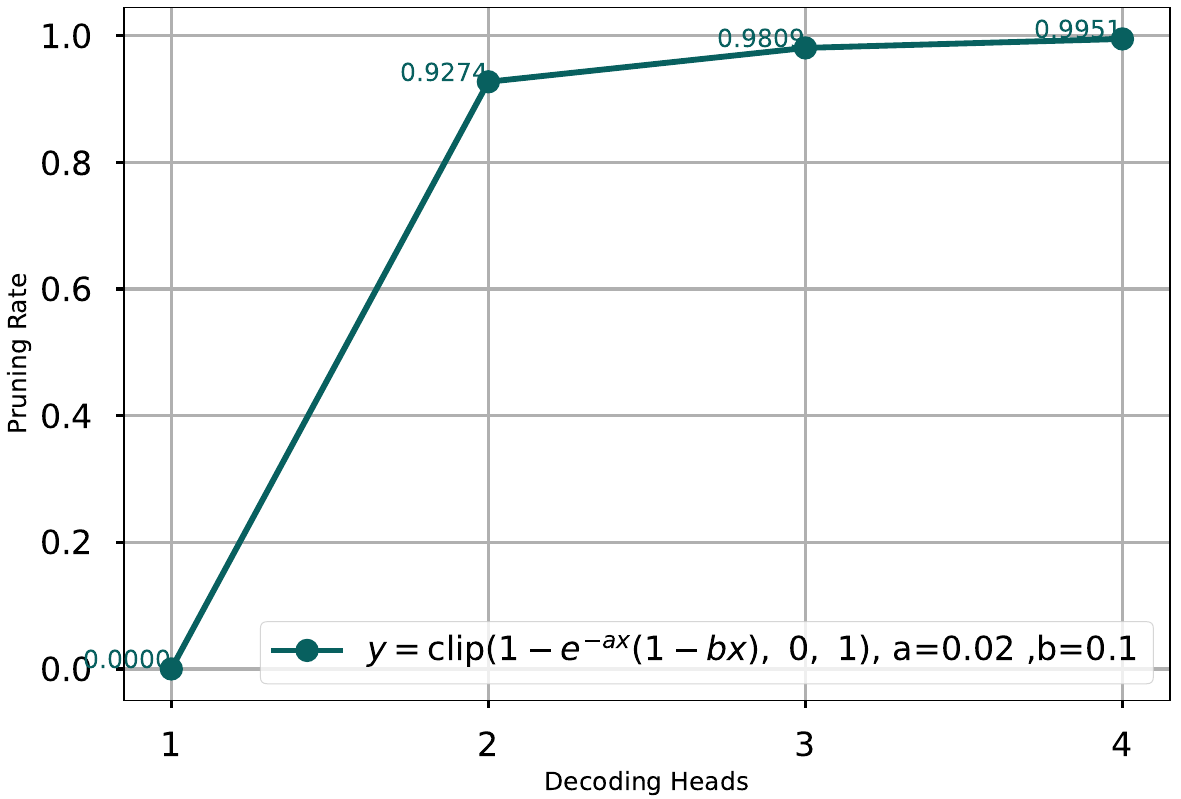}
\caption{Applied pruning rates per-level on a full attention tree with 4 decoding heads.}
\label{fig:prunerate}
\end{minipage}
\centering
\begin{minipage}{0.8\linewidth}
\includegraphics[width=\linewidth]{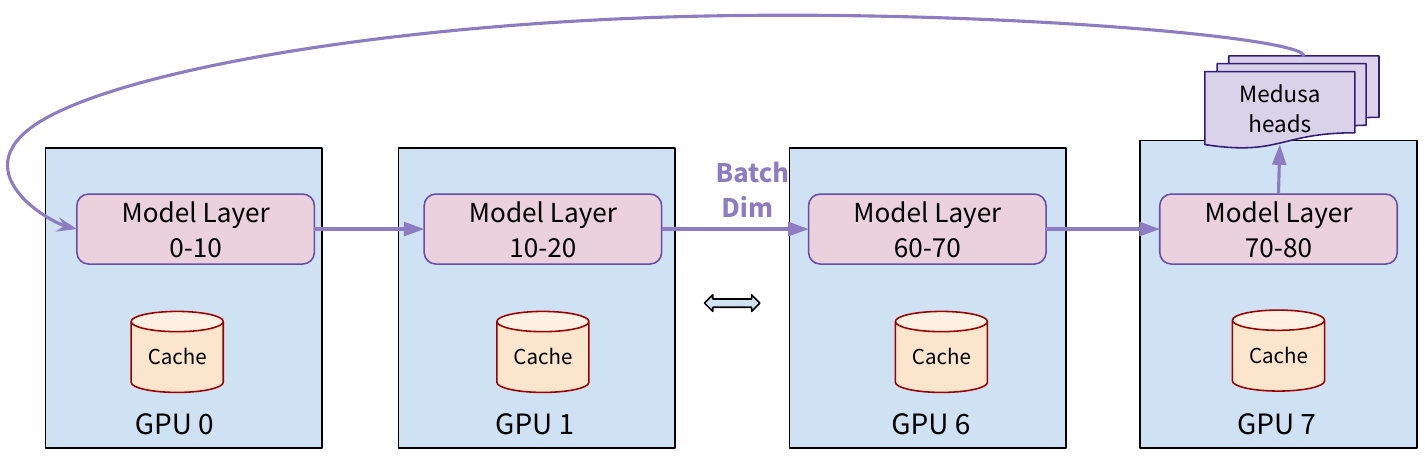}
\end{minipage}
\caption{Distributed LLama-2-70B\cite{llama70b,llama2} on 8 AMD MI250 GPUs\cite{amd_mi250}, where base model and decoding heads are shared by multiple batches. KV cache is also distributed for model layer splits.}
\label{fig:distr}
\end{figure}

\subsection{Optimizing attention mask} \label{chapter-tree}
We study the best-performing attention mask structure and notice that the best practice of achieving higher acceptance rates is retaining all arity ($k$) tokens on the first level of the tree, which fully includes the top-$k$ tokens of the first decoding head. This approach helps enable better parallelism since the subsequent verification of a continuation is cumulative. Therefore, we introduce an arity variable and assume a default tree budget of minimum $k$ (arity) $*$ $l$ (heads/levels) nodes. On smaller devices where such minimal tree mask cannot be met by budgets, SpecMemo allows space for the mask by automatically manipulating other inference hyper-parameters as given in Algorithm \ref{algo1}. For building an attention mask representation, we support two modes of tree construction: 

\paragraph{Custom tree build via pruning:} Given an attention mask configuration with levels ($l$) and arity ($k$), SpecMemo computes pruning rates to target a budget-friendly and realistic size for the new mask, by following the function in Figure \ref{fig:prunefunc}. SpecMemo employs the illustrated scaled logistic function to compute the remaining nodes on an assumed full attention tree. If an existing attention mask is passed, SpecMemo performs the pruning in-place by applying similar ratios, as given in in Figure \ref{fig:prunerate}. The scaled logistic function keeps pruning rates low at first levels of the tree to respect logit variety, and increase the rate on deeper levels to avoid growing a too large tree. For in-place pruning strategy on existing masks, we adopt and refine Medusa’s\cite{cai2024medusasimplellminference} right-to-left pruning strategy on each level, which favors the retention of individual high-probability tokens during top-k sampling from heads. 

\paragraph{Custom tree build from tree features:} SpecMemo derives budget-friendly and computationally feasible tree configurations as a pair of total node and leaf node counts (representing candidate sequences on the tree), as shown in Algorithm \ref{algo1}. As a means of more strict memory management on small GPUs, SpecMemo directly builds tree mask structures with exact features rather than approximating a tree mask size via pruning.

\subsection{Distributed and batched speculative decoding} \label{chapter:batch}
\begin{figure}[htbp]
\centering
\begin{minipage}{0.48\linewidth}
\centering
    \includegraphics[width=\linewidth]{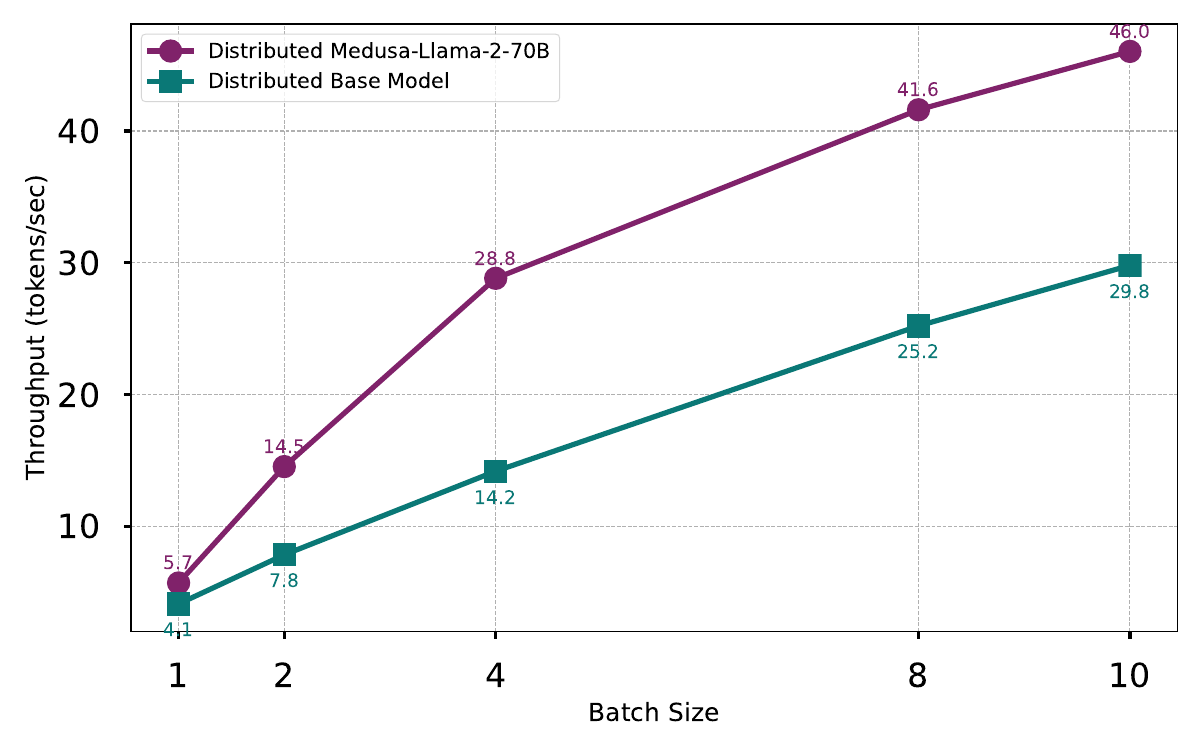}
     \caption{8x throughput gain of distributed and batched Llama-2-70B\cite{llama70b,llama2} on MT-Bench\cite{Bai_2024,mtbench} and 8 AMD MI250\cite{amd_mi250}. 2x improvement over the distributed and batched vanilla decoding.}
     \label{fig:distributed}
\end{minipage}
\hfill
\raisebox{0.001cm}{ 
\begin{minipage}{0.48\linewidth}
\centering
\includegraphics[width=\linewidth]{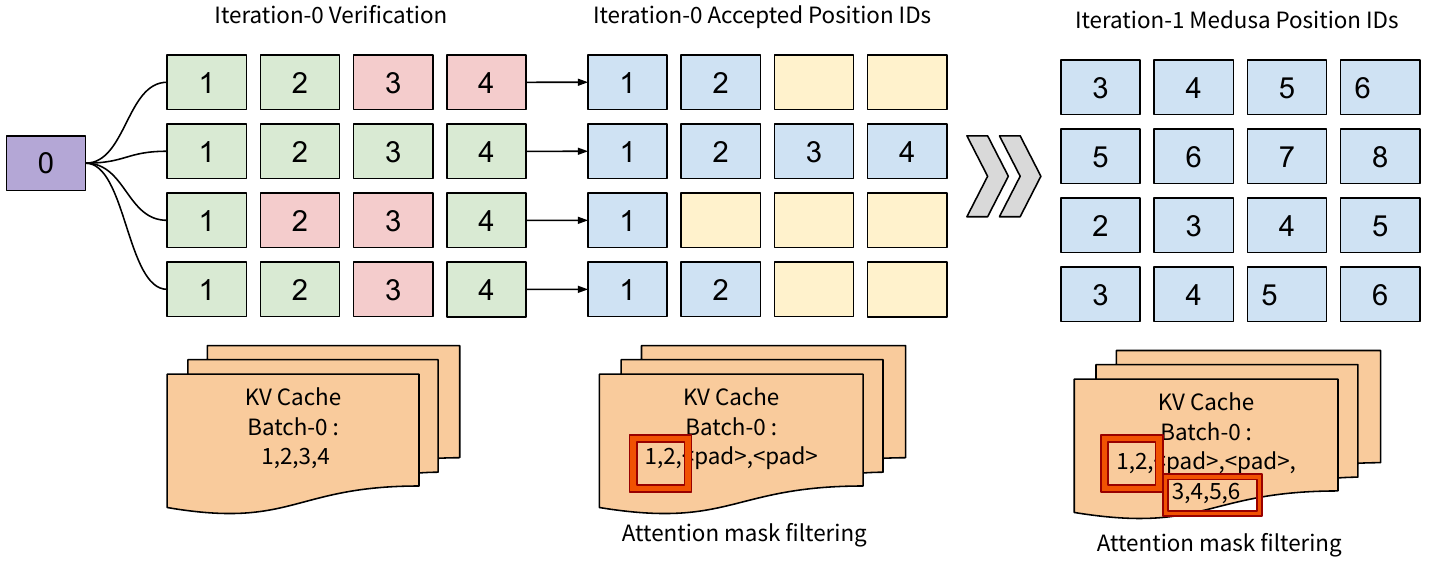}
\caption{A toy example of batched Medusa\cite{cai2024medusasimplellminference}, where attention masks filter pad tokens in the cache. Iteration verification determines whether to discard or keep tokens. Batches have varying length of acceptance; 2,4,1, and 2, respectively. Uniform tensor shape is obtained via padding after verification.}
\label{fig:batch}
\end{minipage}
}
\end{figure}

We distributed the base model evenly by partitioning its layers into equal-sized chunks across all available GPUs, with each GPU also hosting the corresponding slices of the KV cache alongside the layers, given in Figure \ref{fig:distr}. We implement a simple batched speculative decoding method that avoids the need for additional pointer management over padded tokens. The following steps outline in-place modifications made to the speculative decoding for implementing our batching approach. For this algorithm, an example is illustrated on Figure \ref{fig:batch}.
\begin{itemize}[leftmargin=0.5cm, noitemsep, topsep=0pt,labelsep=0.5em]
\item In each batch sequence, candidate tokens are discarded after token rejection in a cumulative way. Input IDs for the accepted tokens across batch dimension are padded to a fixed-length sequences to form uniform tensors for base model verification. 
\item Each batch maps Input IDs to Position IDs in a way to ignore the pad tokens, so that positional embeddings continue from the latest sequence length at the previous iteration.
\item Before attention computation, cached tokens are searched for pad tokens and the corresponding locations in the attention mask are set to $-\infty$. This helps maintain the accuracy of the attention mechanism.
\end{itemize}

\subsection{Algorithm}
The key advantage of SpecMemo is its ability to target successful speculative text generation on mobile GPUs with as little memory as 8GB, by using an adaptable device memory limit in budget optimizer. SpecMemo inference engine follows the heuristic-based Algorithm \ref{algo1} to promise a good performance while conforming to limits of the underlying architecture, which is illustrated on Figure \ref{fig:specmemo}. First, it performs a precise KV cache allocation and prioritizes minimizing the number of decoding heads over coarse-grained memory reductions like model quantization. SpecMemo can be easily integrated into state-of-the-art speculative decoding algorithms, as shown for Medusa\cite{cai2024medusasimplellminference} in Algorithm \ref{algo2}.

\begin{figure}[h!]
\scalebox{0.49}{
\begin{minipage}[c]{\linewidth}
    \begin{algorithm}[H]
    \caption{SpecMemo Algorithm}
    \begin{algorithmic}[1]
    \Function{OptimizerEngine}{$query\_count$, $query\_length$}
        \State $num\_heads \gets default\_heads$
        \State $default\_tree \gets (64\_nodes, 42\_sequences)$
        \State $pruned\_tree\_configs \gets \emptyset , attention\_trees \gets \emptyset$
        \State $min\_cache \gets \Call{ComputeMinCache}{query\_count, query\_length}$
        \State $free\_memory \gets \Call{AvailMemory}{max\_memory, min\_cache, default\_tree, model}$
        \If{$free\_memory$}
            \State $default\_config \gets {min\_cache, default\_heads, default\_tree}$
            \State \Return \Call{Medusa}{default\_tree, default\_config, model}
        \EndIf
        \While{$pruned\_tree\_configs = \emptyset$}
            \State $pruned\_tree\_configs \gets \Call{ExploreTree}{max\_memory, min\_cache, num\_heads}$
            \ForAll{$pruned\_tree\_config \in  pruned\_tree\_configs $}
                \State $total\_tree\_nodes,total\_leaf\_nodes \gets pruned\_tree\_config.properties$
                \State $tree\_mask \gets \Call{BuildCustomTree}{total\_tree\_nodes, total\_leaf\_nodes}$
                \State $attention\_trees \gets attention\_trees \cup \{tree\_mask\}$
            \EndFor
            \If{$attention\_trees != \emptyset$ }
                \State \Return \Call{Medusa}{attention\_trees, pruned\_tree\_configs, model}
            \EndIf
            \State $new\_heads \gets num\_heads$
            \For{$num\_heads \gets num\_heads - 1$ \textbf{to} $2$}
                \If{$\Call{AvailMemory}{min\_cache, num\_heads, default\_tree}$}
                    \State $new\_heads \gets num\_heads$
                    \State \textbf{break}
                \EndIf
            \EndFor
            \If{$new\_heads != num\_heads$}
                \State $model \gets \Call{QuantizeBaseModel}{model, new\_precision}$
            \EndIf
            \State $pruned\_tree\_configs \gets \emptyset$
            \State $attention\_trees \gets \emptyset$
        \EndWhile
    \EndFunction
    \end{algorithmic}
    \label{algo1}
    \end{algorithm}
\end{minipage}
}
\hfill
\begin{minipage}{0.49\linewidth}
 \begin{tikzpicture} [scale=0.71]
    \begin{axis}[%
        view={60}{30}, 
        xlabel style={rotate=-35},
        ylabel style={rotate=8},
        xlabel={KV Cache (GB)}, ylabel={Attention Mask Size (Tree Nodes)}, zlabel={\#  Decoding Heads}, 
        grid=major, 
        zmin=1, zmax=5,ymin=1,xmin=0, ymax=70,
        ytick={5, 16, 27, 31,44,54,64},
        yticklabels={5, 16, 27, 31,44,54,64},
        xtick={1, 3, 5, 10},
        xticklabels={1, 5.2, 7.2, 10},ztick={2,3,4,5},
        colormap/cool,  colorbar,  colorbar style={
            ylabel={Per-Token Latency (msec)}, 
            y label style={at={(axis description cs:-0.15,.5)}, anchor=south},
            ylabel style={font=\tiny}, 
            width=5pt,  
        },
        point meta min=15, 
        point meta max=30, 
    ]
    \addplot3 [
        surf, 
        shader=interp, 
        point meta=explicit, 
        mesh/rows=7, 
        mesh/cols=7, 
        mesh/check=false,
    ] table [meta index=3]{numhead2.dat};

    \addplot3 [
        surf, 
        shader=interp, 
        point meta=explicit, 
        mesh/rows=7, 
        mesh/cols=7, 
        mesh/check=false,
    ] table [meta index=3]{numhead3.dat};

    \addplot3 [
        surf, 
        shader=interp, 
        point meta=explicit, 
        mesh/rows=7, 
        mesh/cols=7, 
        mesh/check=false,
    ] table [meta index=3]{numhead4.dat};

    \addplot3 [
        only marks,
        scatter, 
        mark=x,
        mark options={draw=yellow, line width=1.5pt},
        mark size = 3pt,
        scatter src=explicit,
        scatter/use mapped color={draw=black}, 
    ] table [meta index=3] {numhead5.dat};

    \addplot3 [
        surf, 
        shader=interp, 
        point meta=explicit, 
        mesh/rows=7, 
        mesh/cols=7, 
        mesh/check=false,
    ] table [meta index=3]{numhead5.dat};
    \end{axis}
\end{tikzpicture}
\caption{Token latency exploration of Vicuna-7B\cite{vicuna7b} while varying inference hyper-parameters and mask size on story telling benchmark. Using 3 heads with mask of size 44 gives the best latency.}
\label{fig:maskmodel}
\end{minipage}
\end{figure}

\section{Evaluation}
\begin{figure}[h!]
    \centering
    \begin{minipage}[t]{0.49\linewidth}
        \includegraphics[width=\linewidth, height=7cm]{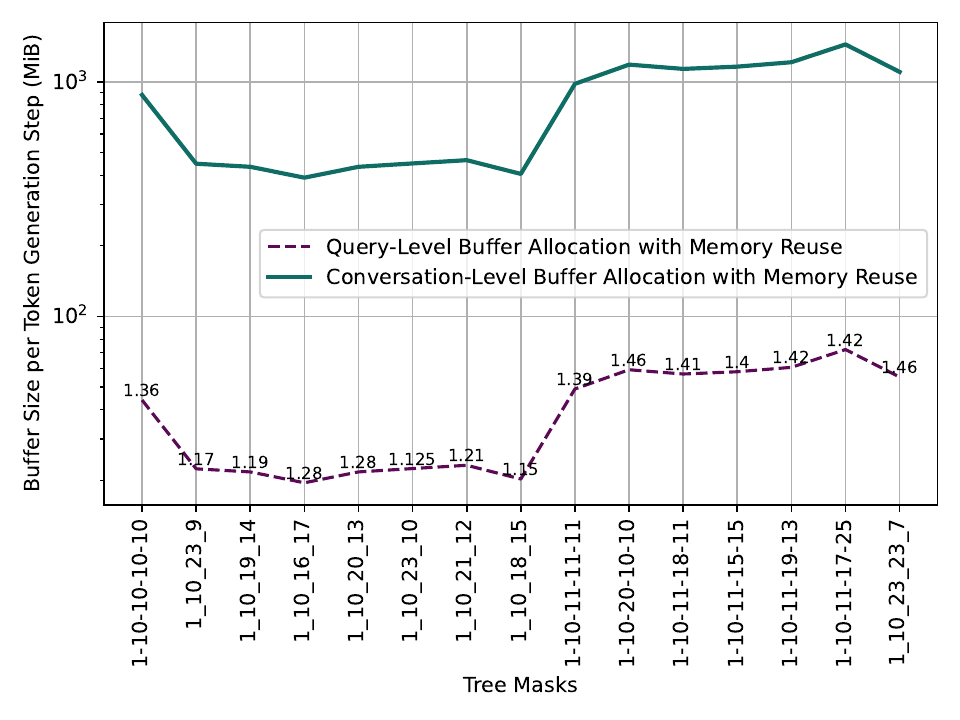}
        \caption{Explored tree-based attention masks with their per-level node counts given in mask labels, i.e, 1-10-19-14 nodes. Annotated numbers on query-level allocation show the average acceptance length ($\tau$) of corresponding masks.}
        \label{fig:maskmem}
    \end{minipage}
    \hfill
    \hfill
    \begin{minipage}[t]{0.49\linewidth}
        \raisebox{1cm}{
            \includegraphics[width=\linewidth, height=6cm]{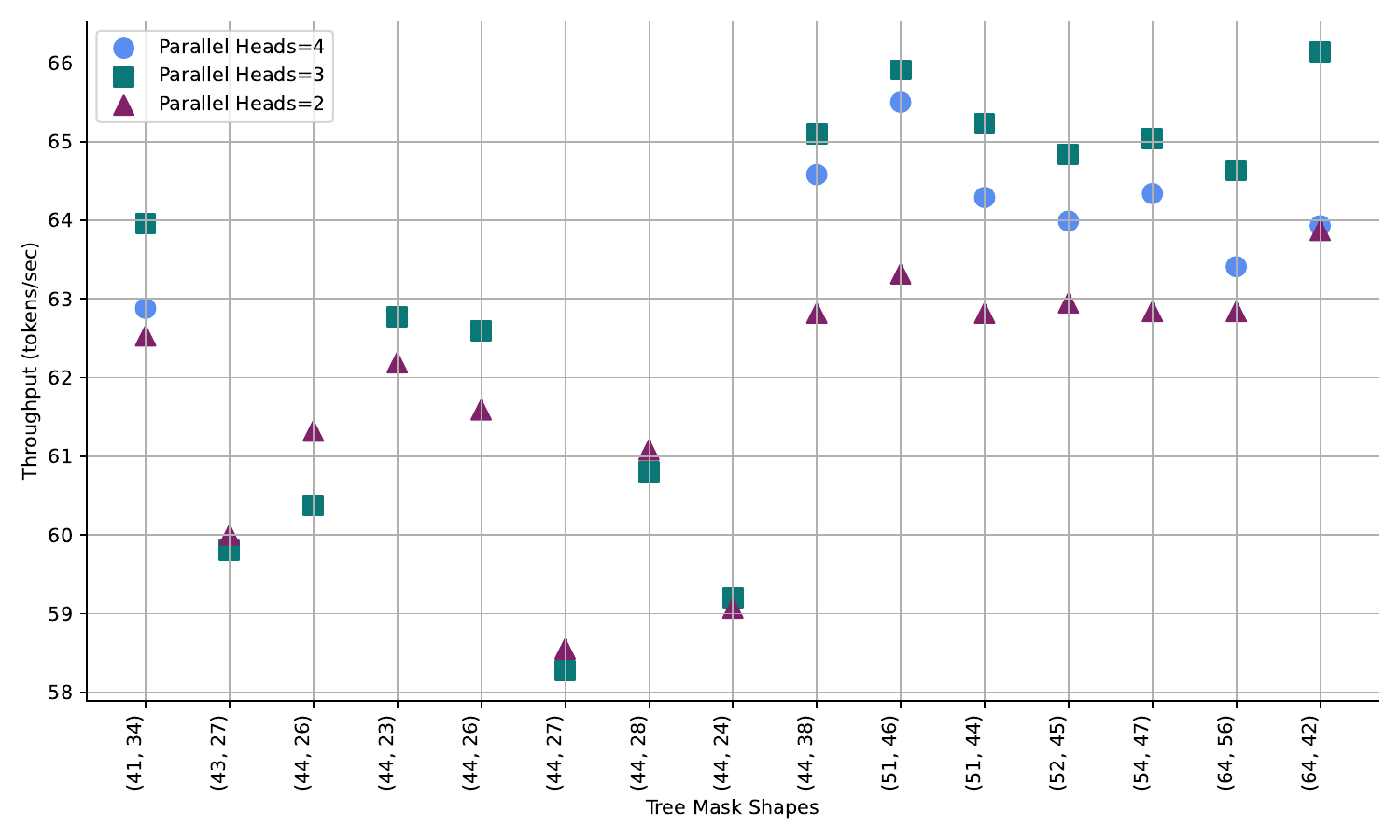}
        }
     \caption{Tree masks used in Figure \ref{fig:maskmem} are respectively labeled with their shapes in (\#nodes, \#leaves) format. Performance of generation on MT-Bench\cite{mtbench} per mask attention mask on single Nvidia Titan RTX with 24GB\cite{nvidia_titan}. }
    \label{fig:maskthroughput}
      \end{minipage}
\end{figure}

On single constrained GPUs, with both tree construction methods of SpecMemo, we observe highly retained throughput gains and improved per-token latency relative to larger set of masks, which empirically aligns with the theoretical analysis on memory-constrained setups presented in Section \ref{chapter:3.1}. In Appendix, we present the statistics on generated attention mask structures on Table \ref{tab:treefeatures}. Figure \ref{fig:maskmodel} presents improved per-token latency relative to larger set of masks and performance impact of varying hyper-parameters with SpecMemo. In particular, in a multi-turn dialogue setting where a tree mask is repeatedly applied across multiple query generation phases, smaller masks crafted by SpecMemo highly contributes to overall memory efficiency savings on small GPUs. Figure \ref{fig:maskmem} shows the gap of memory allocations between query generation and entire chatbot completion for explored attention masks. Tree mask built by SpecMemo named 1-10-16-17 in Figure \ref{fig:maskmem} occupies 19.5 MB buffer per query, while allocating 390 MB across 20 queries in MT-Bench\cite{mtbench}. Compared to the last mask (provided by Medusa \cite{cai2024medusasimplellminference}) with 55 MB buffer allocation per query, which accumulates 1.1 GB over the conversation, this mask delivers 65\% runtime memory reduction in a chat, while retaining 96\% throughput (Figure \ref{fig:maskthroughput}.

\begin{figure}
\centering 
\begin{minipage}{0.8\linewidth}
 \includegraphics[width=\linewidth]{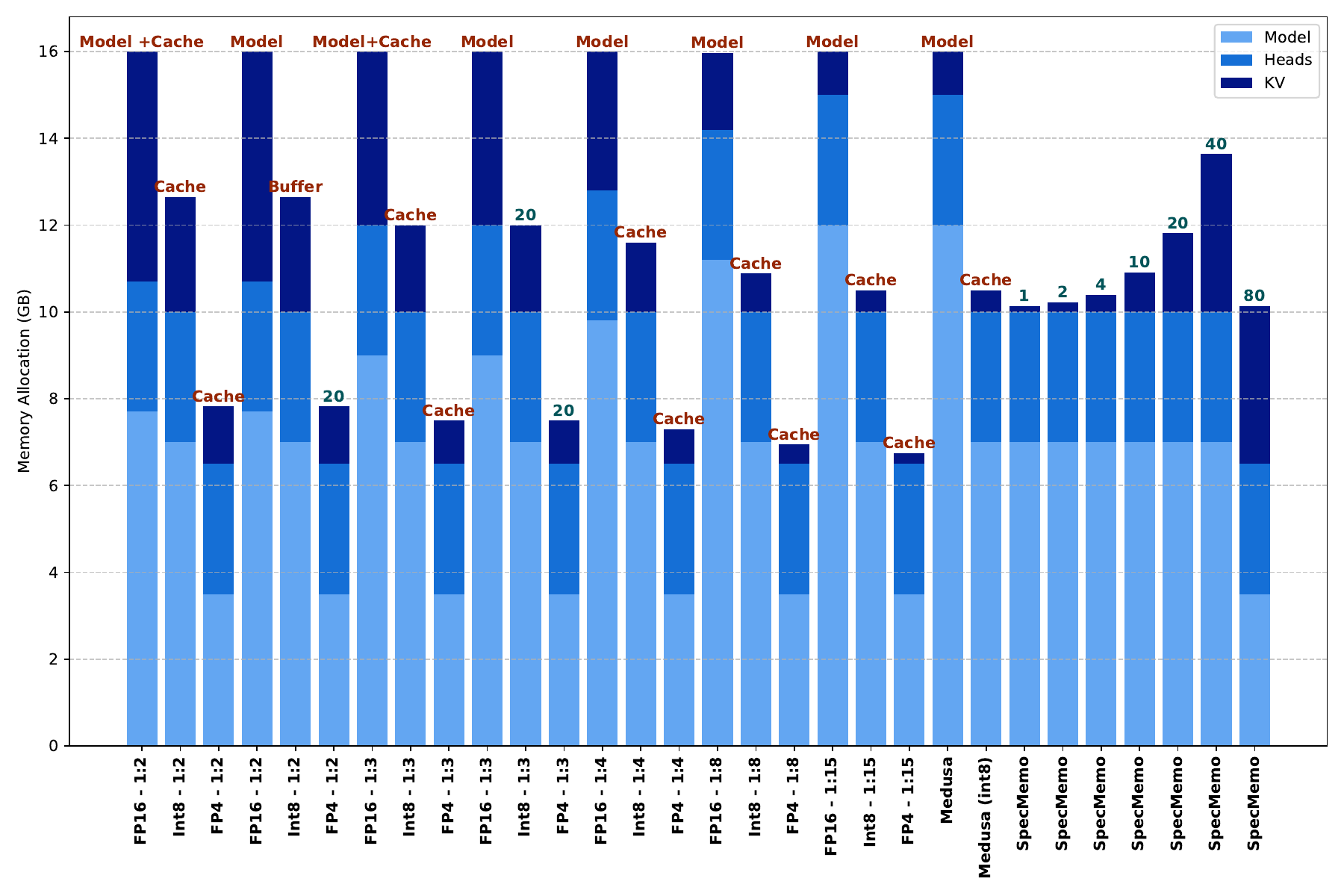}
\caption{Comparison of SpecMemo against memory scaling ratios ranging from 1:2 to 1:15, where each x:y ratio denotes allocating memory first to the KV cache followed by the model, conducted on an AMD AI X1 Mini Pro\cite{amd_x1pro} mobile device with a 16 GB GPU\cite{amd_radeon}. Failed configurations are annotated with the corresponding OOM reason (Model/Cache/Buffer). Successful cases are annotated with the total number of queries run. }
\label{fig:baseline}
\vspace{-0.4cm}
\end{minipage}
\end{figure}

 Furthermore, we compare SpecMemo budget allocator against a memory scaling scheme built on top of the following memory management strategy: 
\begin{itemize}[nosep, leftmargin=0.5cm, labelsep=0.3em]
\item To run queries successfully, one needs to decompose the device memory among KV cache, base model and decoding head allocations, which accounts for the majority of constrained GPU memory. We use memory ratios as our baseline to distribute available memory to speculative decoding memory requirements. Buffer memory is the smallest allocation, therefore the scaling can be simplified as the fraction of cache allocation to remaining, model-based memory allocations. For example, on a device with 16 GB VRAM, allocating 10-15 GB for loading models and reserving 1-5 GB for cache is standard memory distribution required by successful inference. On a GPU with 16 GB VRAM, ratio 1:2 maps to 5.3 GB KV cache preallocation and 10.6 GB model allocations.
\item When porting speculative decoding to a larger memory device, such memory scaling helps linearly increase the space for base model. This means when the model size remains the same, larger devices eliminates quantization dependency of loading the model, if required earlier on the smaller device. \item Following this baseline also helps scale cache size in accordance with the growing LLM size. 
\end{itemize}
As outlined above, described memory scaling is an effective baseline strategy in demonstrating how error-prone inference becomes without finer memory management, such as sophisticated budgeting provided by SpecMemo, on constrained environments. We evaluate this baseline against SpecMemo on Figure \ref{fig:baseline}, where we annotate the number of maximum queries (for $m=128$) that can be successfully served in a chatbot under the given memory allocation scheme. 

\begin{itemize}[leftmargin=0.5cm, noitemsep, topsep=0pt,labelsep=0.5em]
    \item While FP16 model is the main memory pressure factor in all scalings, space reserved for KV cache becomes the bounding factor on the number of maximum allowed queries for Int8 and FP4 scenarios. 
    \item The last column demonstrates SpecMemo's ability to automatically quantize the base model during memory budgeting, whereas the first column fails to fully load the model due to hitting the device limit. 
    \item Remarkably, the second last column demonstrates SpecMemo's effective attention mask adjustment capability to satisy the remaining memory budget for runtime allocations, as opposed to the failed case under baseline in the fifth column.  
\end{itemize}

Additionally, we demonstrate the key advantage of distributed and batched speculative decoding solution, which is increasing the usability of multiple small server GPUs that can serve multiple users efficiently in Figure \ref{fig:distributed}.

\section{Limitations and future work}
This work highlights the potential of resource-limited GPUs techniques in making fast text generation more practical for real-world deployment. However, SpecMemo budget-friendly speculative decoding remains restricted by original generational capacilities of base model LLM and trained context limit allowance, which determines the maximum text generation length of a model. We leave further improvements on batched speculative decoding such as continuous batching to reduce pad tokens to future research. Improved and efficient speculative decoding approaches may also cover image generation tasks.

{
\small

}

\appendix

\section{Appendix / supplemental material}
Table \ref{tab:treefeatures} presents the features of attention masks in number of leaf nodes (candidate sequences) $/$ tree nodes format, where they directly corresponds to tree illustrations in Table \ref{tab:treeshapes}. By intuition, as the number of candidate sequences on the tree increases, acceptance length should be correspondingly increase at generation steps due to allowing more token variety. However, as shown in Table \ref{tab:treefeatures}, the first custom tree with toal 44 nodes under 4 heads possesses 37 candidate sequences. In Figure \ref{fig:maskmodel}, the same attention mask with 44 nodes outperforms the default and original Medusa \cite{cai2024medusasimplellminference} attention mask with 64 nodes that was statistically developed for Vicuna 7B\cite{vicuna7b} model. 

Furthermore, since using 1 decoding head along with the base model (root token) results in the same tree structure across various tree figures, we cut the minimum decoding heads to 2 in SpecMemo implementation and performance evaluations.
\begin{figure}[h!]
\begin{table}[H]
    \renewcommand{\arraystretch}{1.5} 
    \centering
     \caption{Attention Mask Features}
    \resizebox{0.6\linewidth}{!}{  
    \begin{tabular}{@{} >{\raggedleft\arraybackslash}p{0.4cm} @{} cccc @{}}

    \multicolumn{1}{c}{} & \textbf{Heads = 1} \(\rightarrow\) & \textbf{Heads = 2} \(\rightarrow\) & \textbf{Heads = 3} \(\rightarrow\) & \textbf{Heads = 4} \\  

    \hline
    \multirow{4}{*}{\makebox[0pt][r]{\rotatebox{90}{\textbf{Pruned M}}}} & 1/2 & 1/3 & 1/4 & 1/5 \\
&
    1/11 & 1/12 & 7/14 & 10/16  \\  
&
    1/11 & 3/23 & 15/26 & 18/27  \\  
&
    1/11 & 5/34 & 17/26 & 20/31  \\  
    \hline

    \multirow{2}{*}{\rotatebox{90}{\textbf{C}}} &
    10/11 & 19/18 & 28/27 & 37/44 \\  
&
    10/11 & 20/18 & 34/35 & 56/64 \\  
    \hline

    \rotatebox{90}{\textbf{M}} &
   10/11 & 25/34 & 39/57 & 42/64 \\  
    \end{tabular}
    } 
\caption{C refers to Custom Trees explored by SpecMemo. M refers to Medusa\cite{cai2024medusasimplellminference} trees. Pruned M refers to further partitioned Medusa\cite{cai2024medusasimplellminference} trees.}
\label{tab:treefeatures}
\end{table}
\centering
\scalebox{0.8}{
\begin{minipage}[c]{\linewidth}
    \begin{algorithm}[H]
    \caption{SpecMemo for Medusa\cite{cai2024medusasimplellminference}}
    \begin{algorithmic}[1]
    \Function{Medusa}{$attention\_trees, pruned\_tree\_configs, model$}
        \State $results \gets \emptyset$
        \ForAll{$config \in pruned\_tree\_configs$}
            \State $\Call{MedusaModel}{model, config, attention_tree}$
            \State $(acceptance\_length, speedup) \gets \Call{MedusaGenerate}{query\_count, query\_length}$
            \State $results \gets (config, acceptance\_length, speedup)$
        \EndFor
        \State \Return $best\_config \gets \Call{Max}{results.speedup}$
    \EndFunction
    \end{algorithmic}
    \label{algo2}
    \end{algorithm}
    \end{minipage}
}
\end{figure}

\subsection{Ablation Study}
\subsubsection{Candidate Sequence Analysis} \label{chapter:appendtree}
In Medusa-based\cite{cai2024medusasimplellminference} speculative decoding with a tree mask, sampled logits are evaluated to pass a certain threshold to be accepted followed by a cumulative product of probabilities within the sequence. The longest candidate sequence that verified its tokens is accepted over other candidate sequences, which wastes the computation and memory due to rejected sequences. In decoding stage of text-generation, forward passes are performed on candidate sequences in between pre- and post-verification stages of speculative decoding. We individually analyze the likelihood of tree branch (candidate sequence) acceptance before and after the blackbox forward layers are applied to the sequences. Then, we propose a static tree mask that can efficiently reduce resource allocation prior to verification.

\paragraph{Pre-Verification}
We observe the probability distribution of candidate sequences on tree-based speculative decoding. On a static tree of 64 nodes and 42 leaves which represents candidate sequences, we normalize the embeddings of sequences and measure pair-wise cosine similarity. Heatmap analysis in Figure \ref{fig:similarty} suggests that  candidate sequences are highly similar and pruning branches might keep generation quality of original speculative decoding. Since all branches exhibit high cosine similarity pre-verification, selecting only one for the entire generation is an effective way to greatly reduce amount of memory allocations. However, this way of pruning is unlikely to yield as fast model answers due to aggressively restricting variety of token combinations. Although highest amount of resource saving comes from employment of minimum number of candidate branches, corresponding acceptance rate and therefore majority of speedup from speculation degrades if relying on a single tree path, which is analogous to chain-based verification \cite{wu2025specrouteradaptiveroutingmultilevel}. 

\begin{figure*}[ht!]
    \centering
    \begin{tabular}{cccc}
    \subcaptionbox{103 generation steps}
       {\includegraphics[width=0.35\textwidth, height=0.12\textheight]{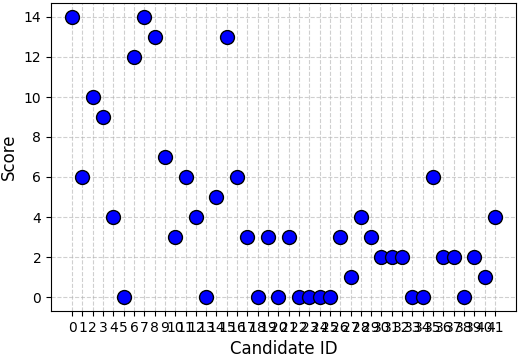}} &
         \subcaptionbox{82 generation steps}
        {\includegraphics[width=0.35\textwidth, height=0.12\textheight]{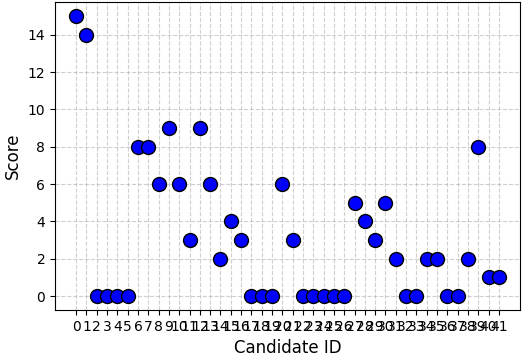}} & \\
        \subcaptionbox{90 generation steps}
       {\includegraphics[width=0.35\textwidth, height=0.12\textheight]{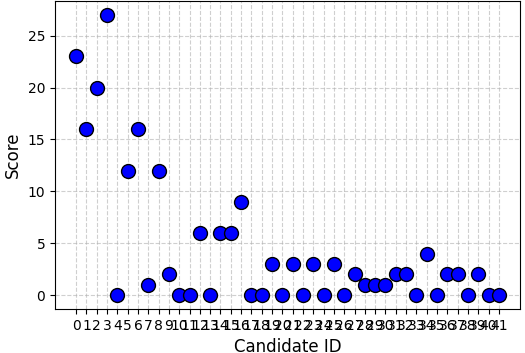}} &
         \subcaptionbox{91 generation steps}
       {\includegraphics[width=0.35\textwidth, height=0.12\textheight]{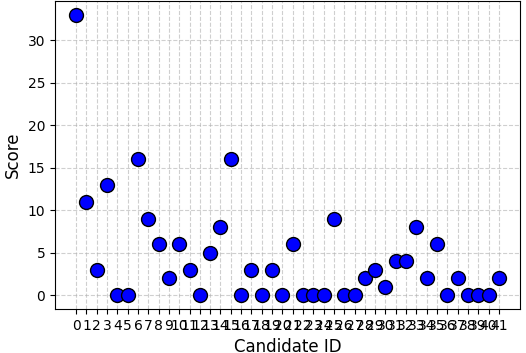}} 
    \end{tabular}
    \caption{Distribution of selected branches and their contribution to total generation length over a continued story generation benchmark with original Medusa\cite{cai2024medusasimplellminference} that forms 42 total candidate branches.}
    \label{fig:branches}
\end{figure*}

\paragraph{Post-Verification}
We analyze the acceptance rate of Medusa\cite{cai2024medusasimplellminference} tree branches that are originally designed to reduce exponential number of speculative candidates generated by parallel speculative heads. As part of our analysis, we profile acceptance rates of each branch to determine their contribution to overall generation length. When selected with a top probability over other branches, any branch can be either partially or fully accepted. Therefore, we assign linear scores of acceptance length to selected branches in each generation step. Since Medusa\cite{cai2024medusasimplellminference} uses typical sampling with top-k probabilities to form the speculative continuations, we expect that leftmost tree branches yield higher importance along with higher speculation scores. Figure \ref{fig:branches} shows that highest sequence length scores are provided by the first one fourth branches over the last three fourth branches of the Medusa\cite{cai2024medusasimplellminference} tree in left-to-right order. 

\begin{minipage}{\linewidth}
\begin{table}[H]
\caption{Attention trees used in the latency experiments given on Figure \ref{fig:maskmodel} }
    \renewcommand{\arraystretch}{1.5} 
    \centering
    \resizebox{\textwidth}{!}{  
    \begin{tabular}{m{1.5cm} cccccc}

    \multicolumn{1}{c}{} & \textbf{Heads = 1} \(\rightarrow\) & \textbf{Heads = 2} \(\rightarrow\) & \textbf{Heads = 3} \(\rightarrow\) & \textbf{Heads = 4} & \textbf{Number of Nodes} \\  

    \hline
    \multirow{4}{*}{\rotatebox{90}{\textbf{Pruned Medusa\cite{cai2024medusasimplellminference} Trees}}} &
    \includegraphics[width=0.18\textwidth]{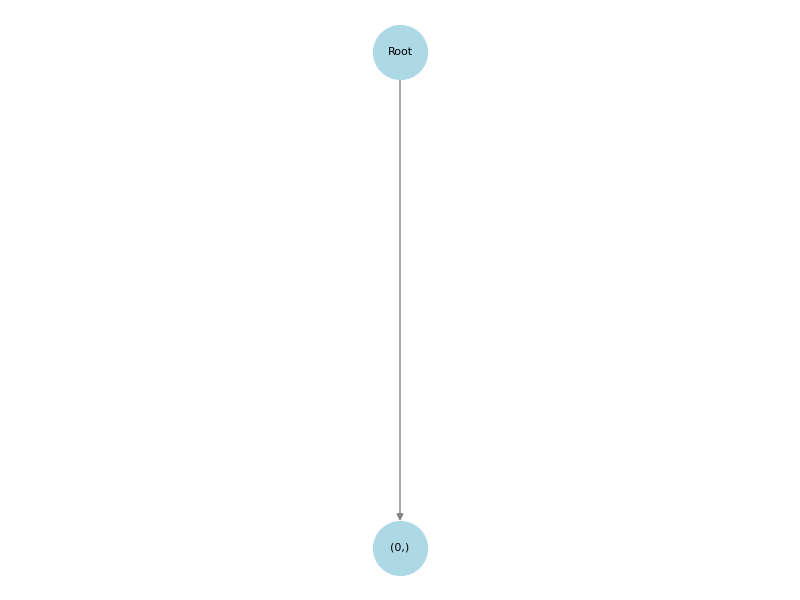} & 
    \includegraphics[width=0.18\textwidth]{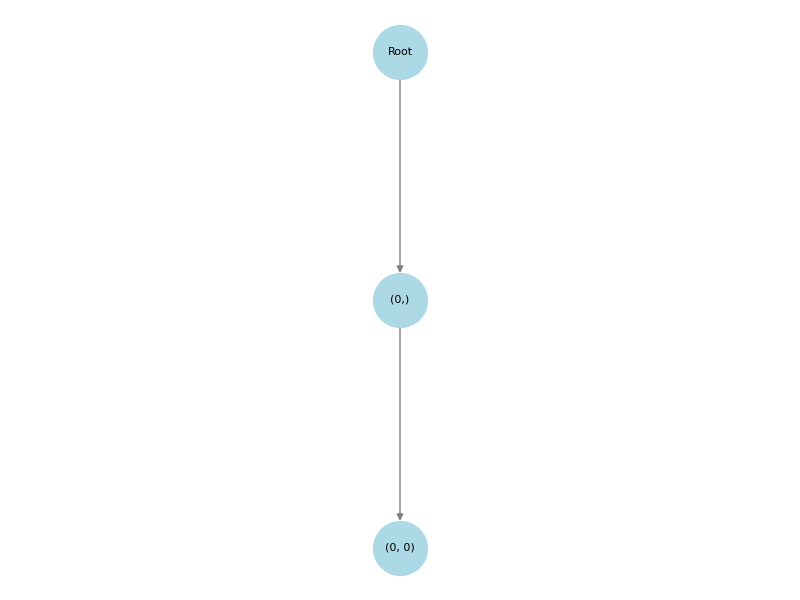} & 
    \includegraphics[width=0.18\textwidth]{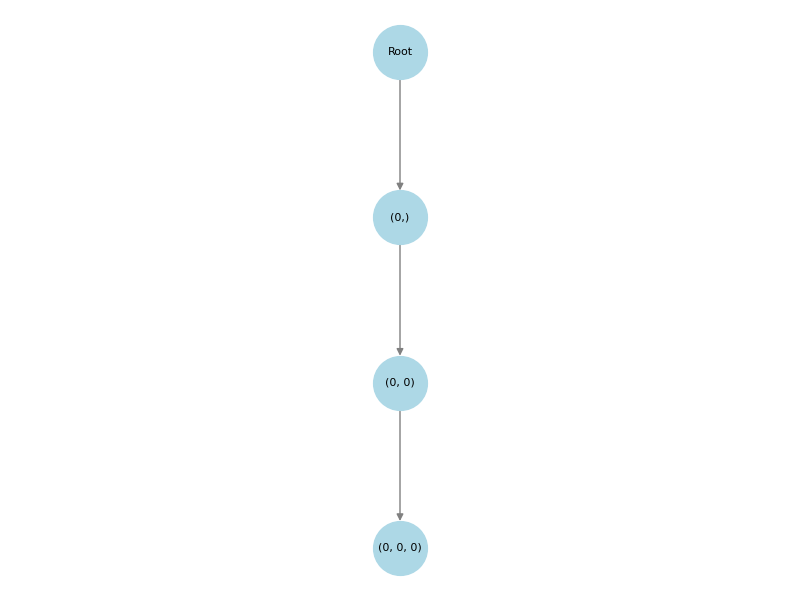} & 
    \includegraphics[width=0.18\textwidth]{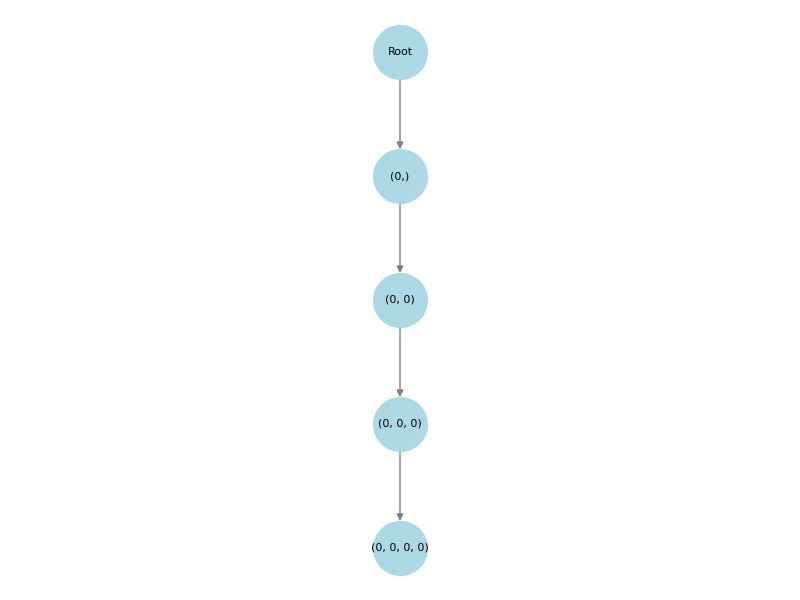} & 5 \\
&
    \includegraphics[width=0.18\textwidth]{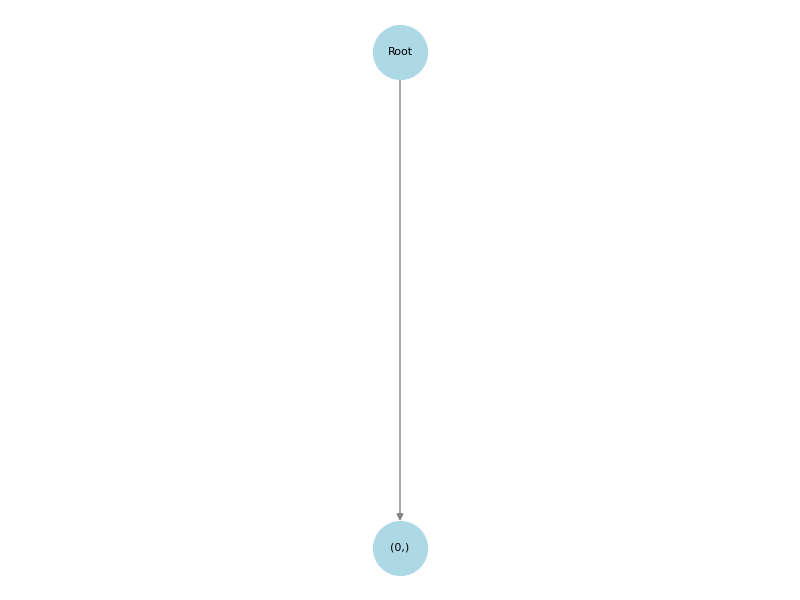} & 
    \includegraphics[width=0.18\textwidth]{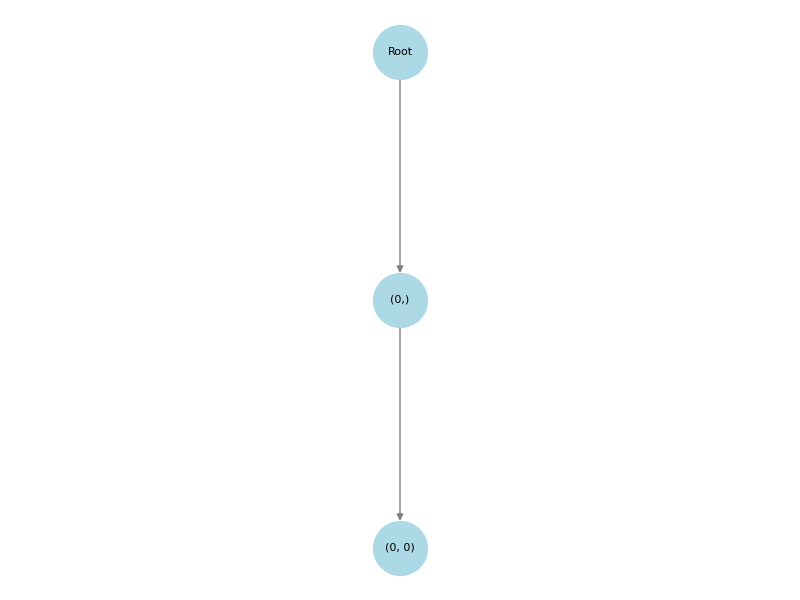} & 
    \includegraphics[width=0.18\textwidth]{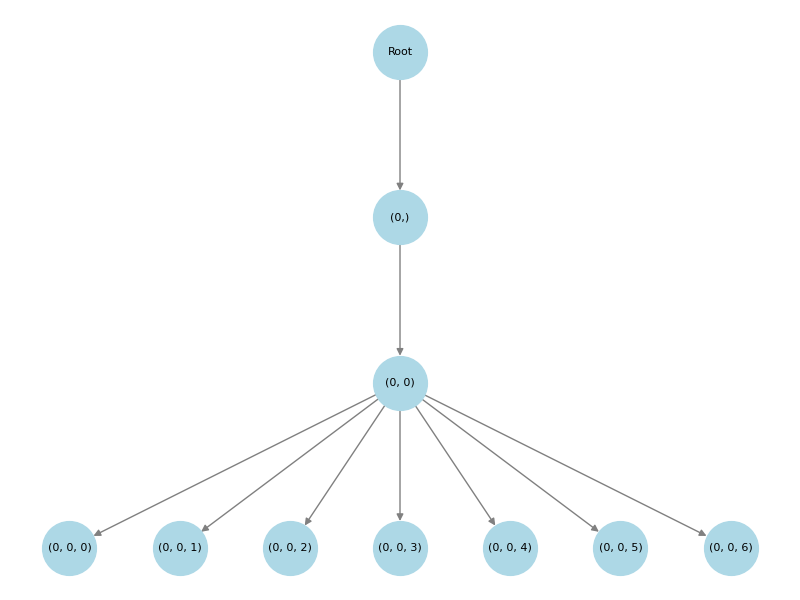} & 
    \includegraphics[width=0.18\textwidth]{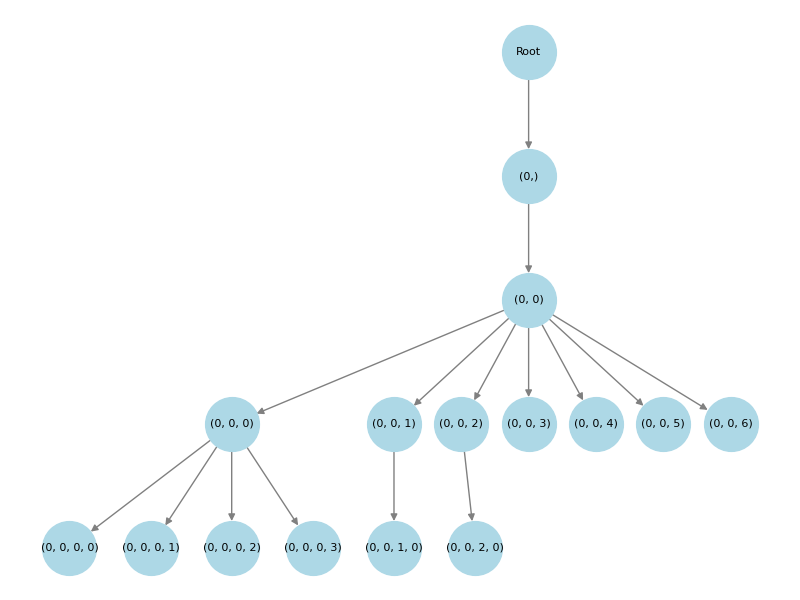} & 16 \\
&
    \includegraphics[width=0.18\textwidth]{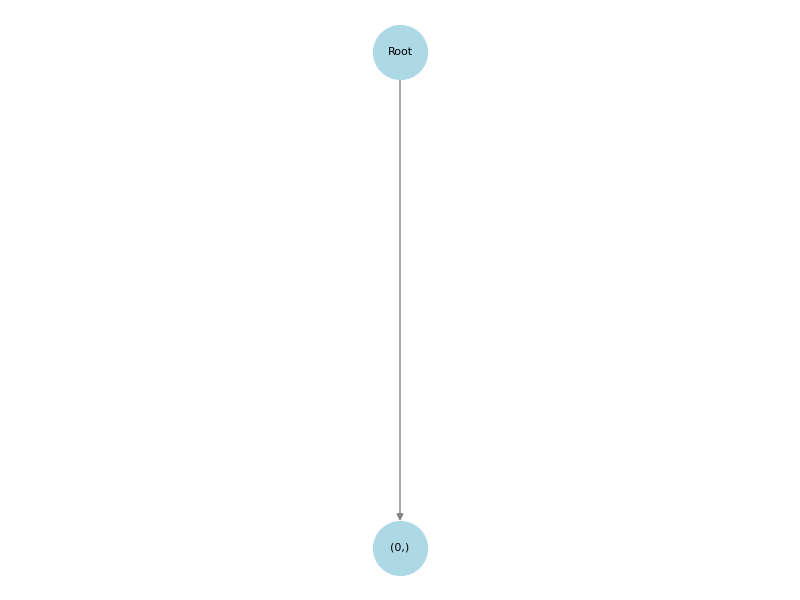} & 
    \includegraphics[width=0.18\textwidth]{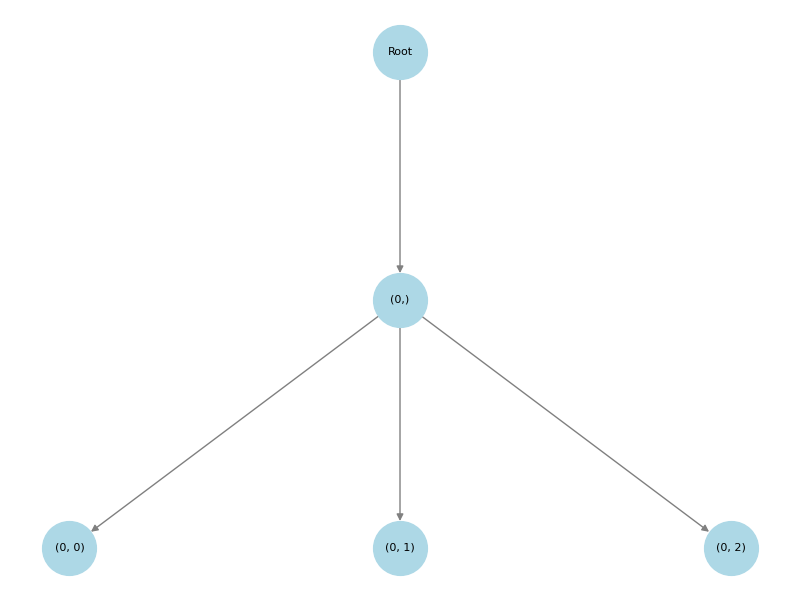} & 
    \includegraphics[width=0.18\textwidth]{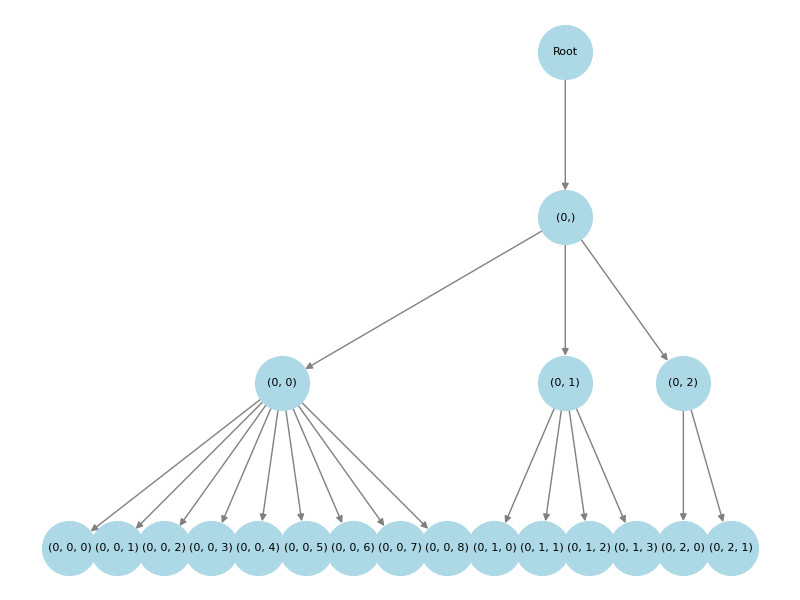} & 
    \includegraphics[width=0.18\textwidth]{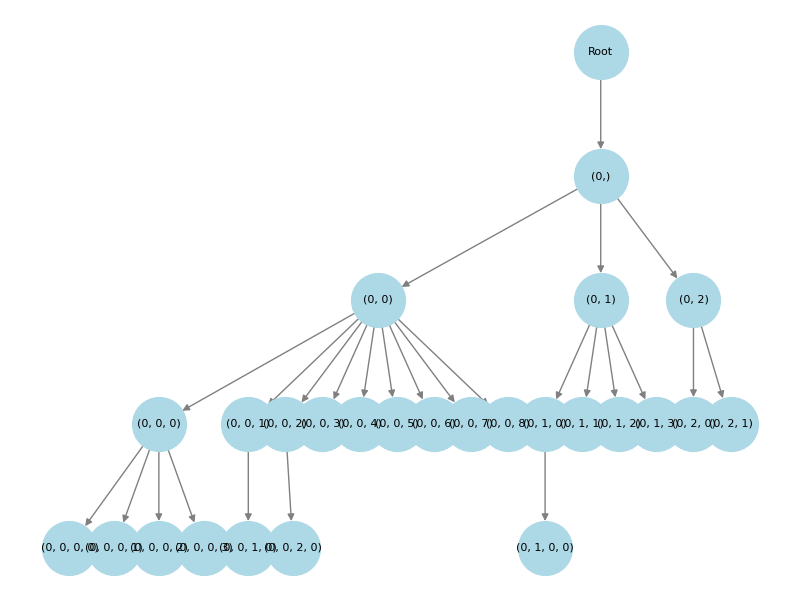} & 27 \\
&
    \includegraphics[width=0.18\textwidth]{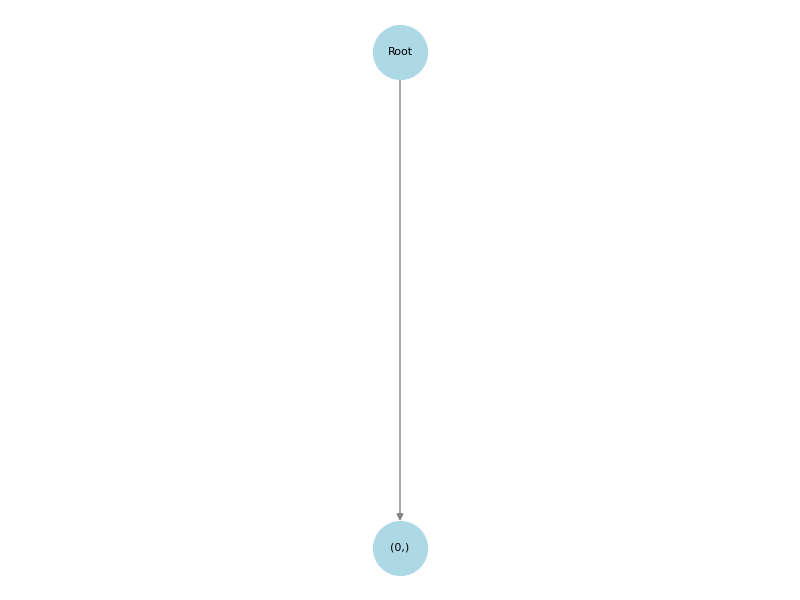} & 
    \includegraphics[width=0.18\textwidth]{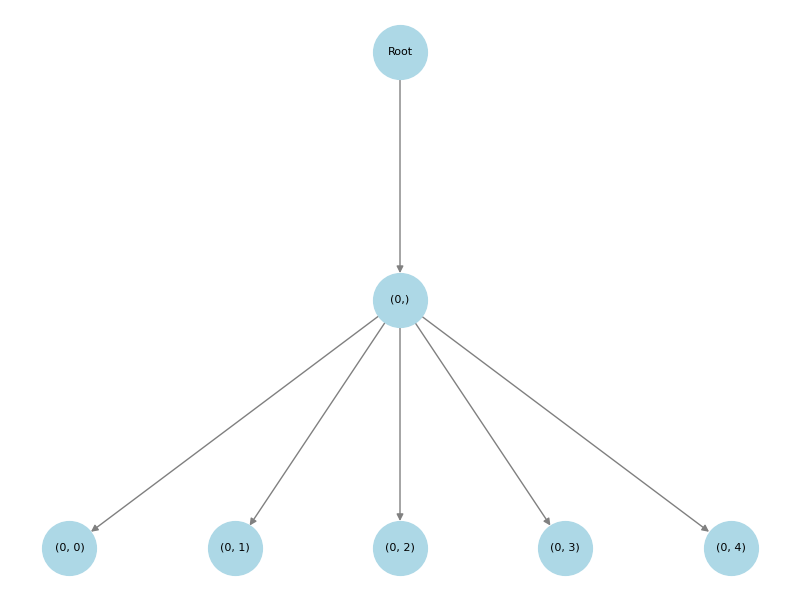} & 
    \includegraphics[width=0.18\textwidth]{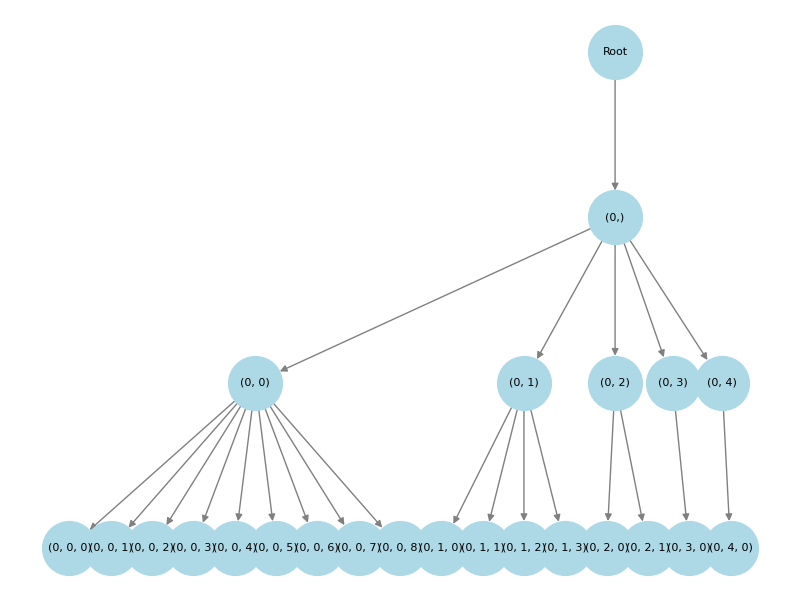} & 
    \includegraphics[width=0.18\textwidth]{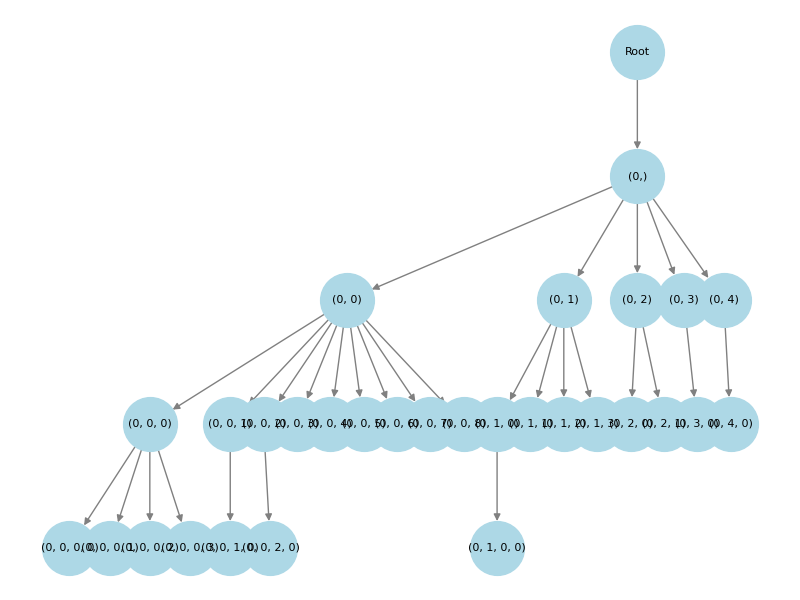} & 31 \\
    \hline
    
    \multirow{2}{*}{\rotatebox{90}{\textbf{Custom Trees}}} &
    \includegraphics[width=0.18\textwidth]{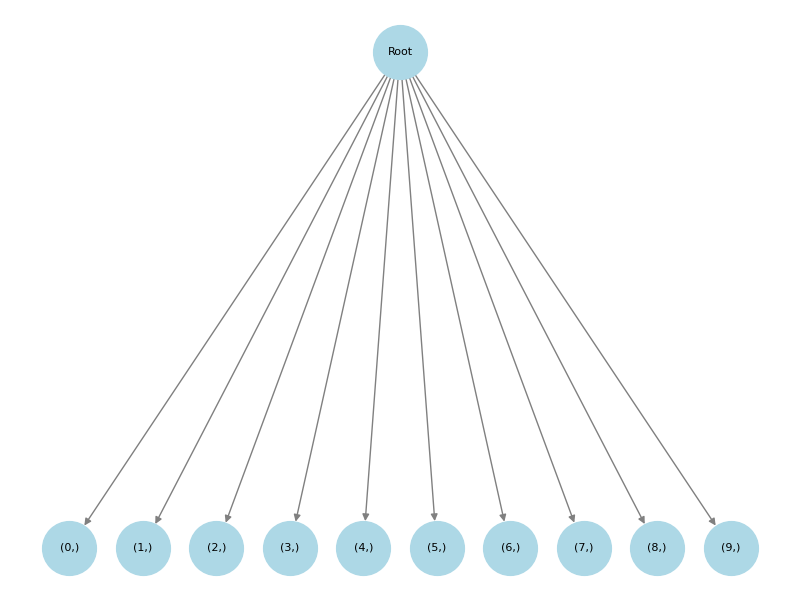} & 
    \includegraphics[width=0.18\textwidth]{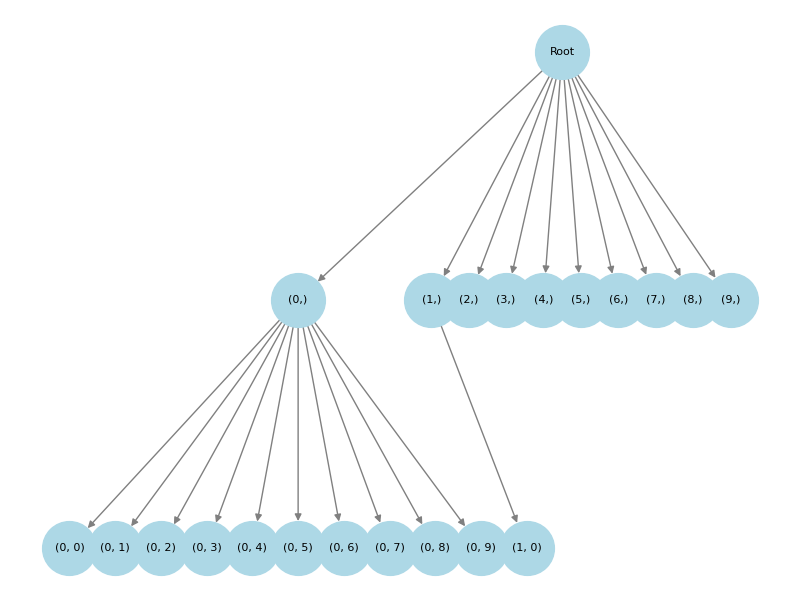} & 
    \includegraphics[width=0.18\textwidth]{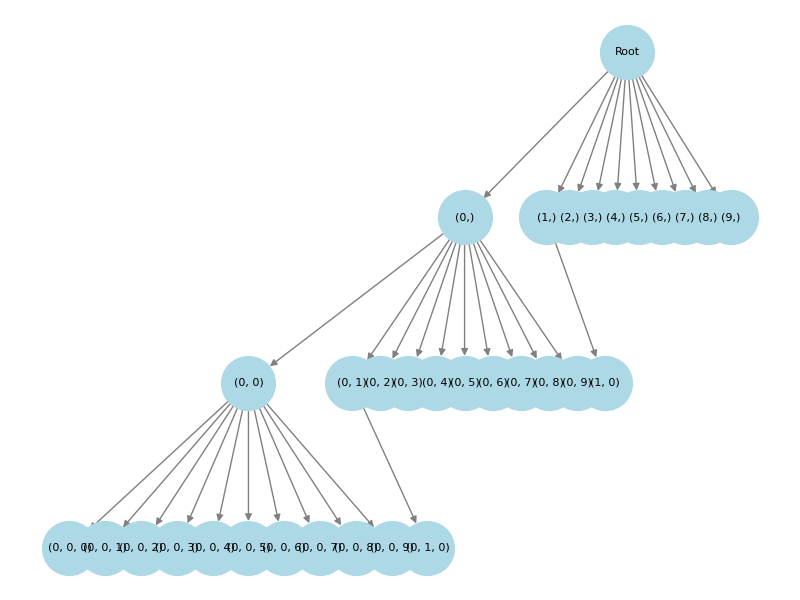} & 
    \includegraphics[width=0.18\textwidth]{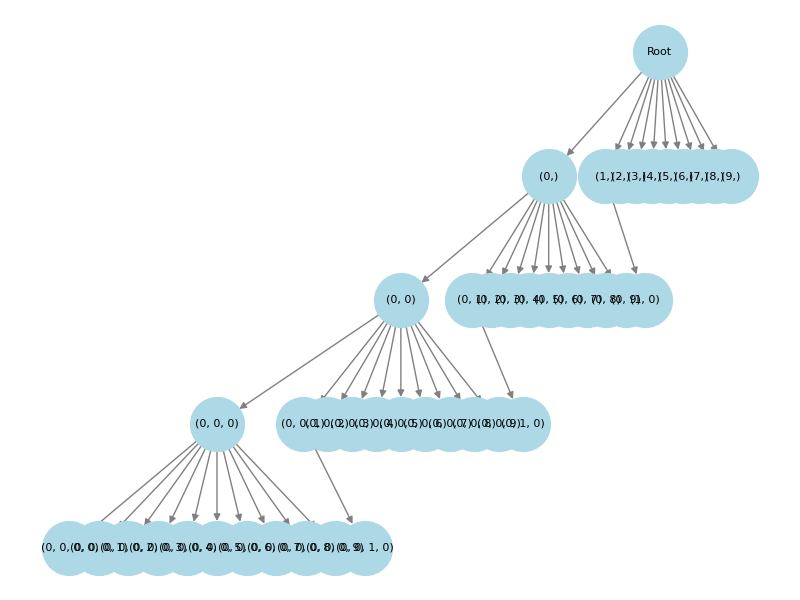} & 44 \\
    &
    \includegraphics[width=0.18\textwidth]{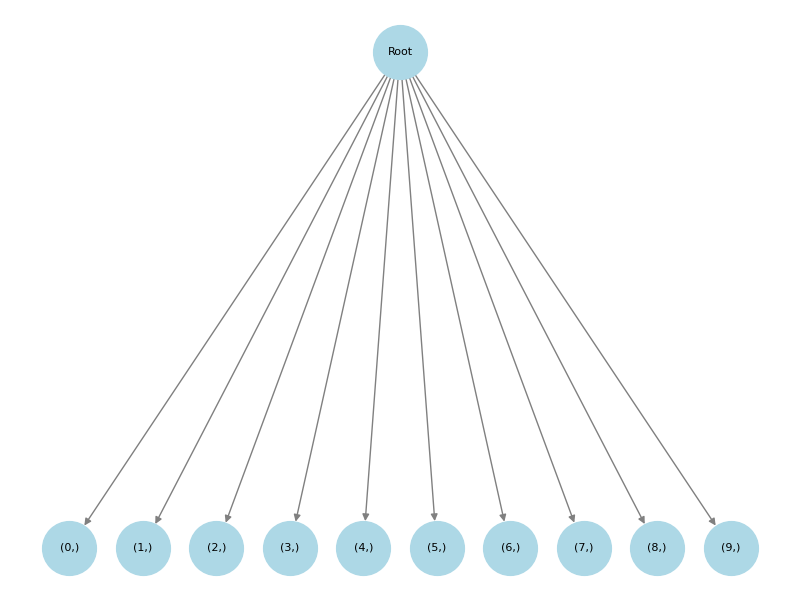} & 
    \includegraphics[width=0.18\textwidth]{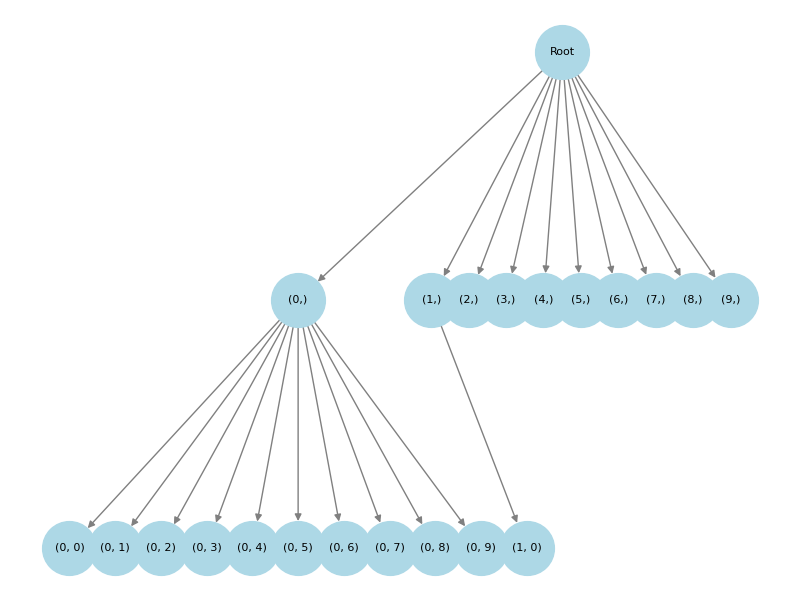} & 
    \includegraphics[width=0.18\textwidth]{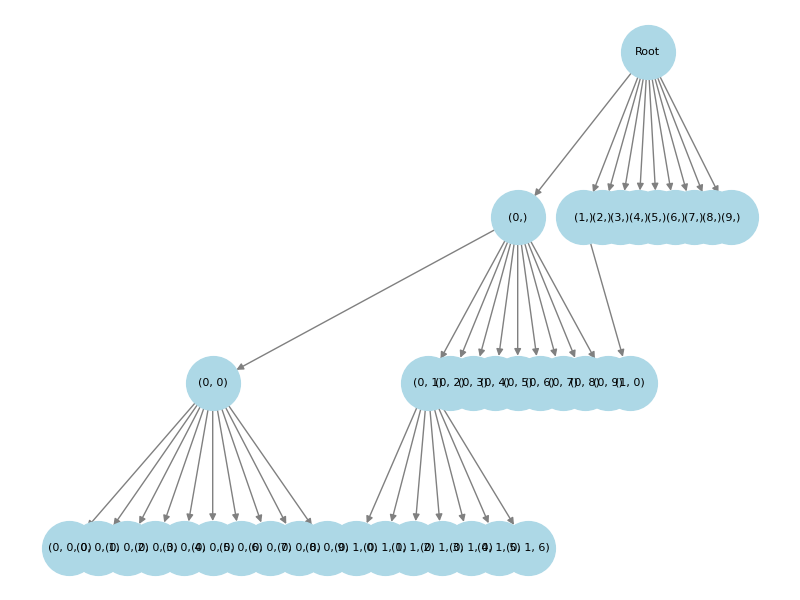} & 
    \includegraphics[width=0.18\textwidth]{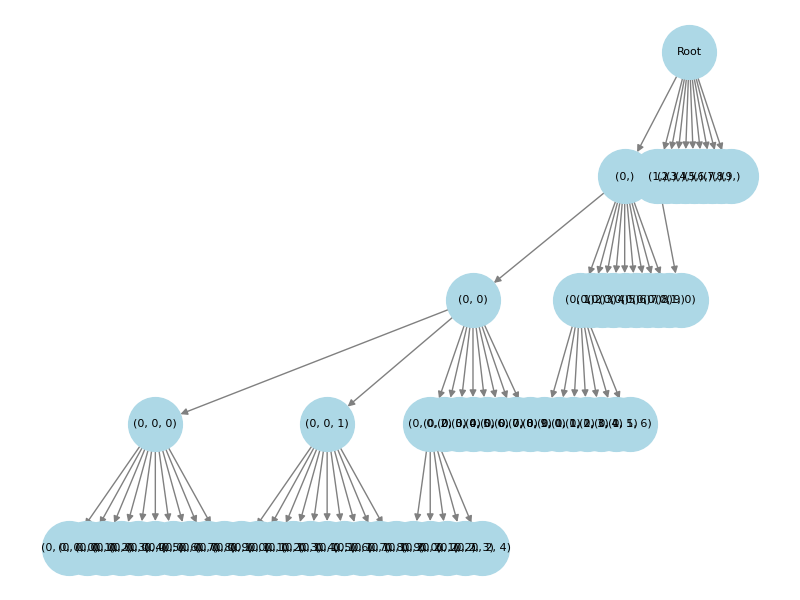} & 64 \\
    \hline
    
    \rotatebox{90}{\textbf{ Medusa\cite{cai2024medusasimplellminference} Tree}} &
    \includegraphics[width=0.18\textwidth]{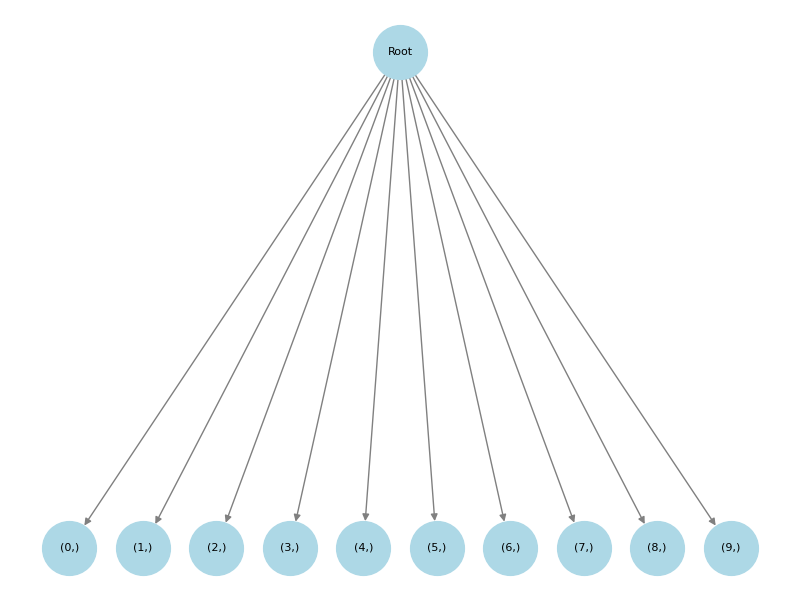} & 
    \includegraphics[width=0.18\textwidth]{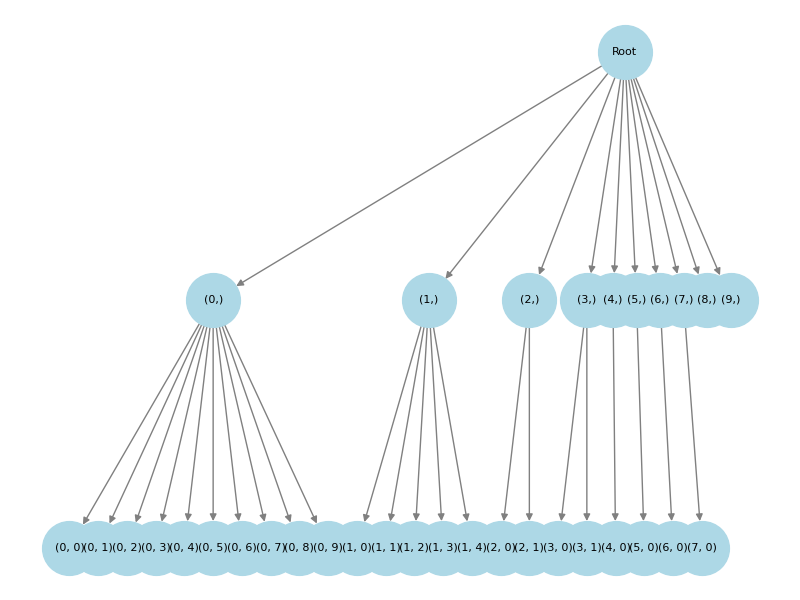} & 
    \includegraphics[width=0.18\textwidth]{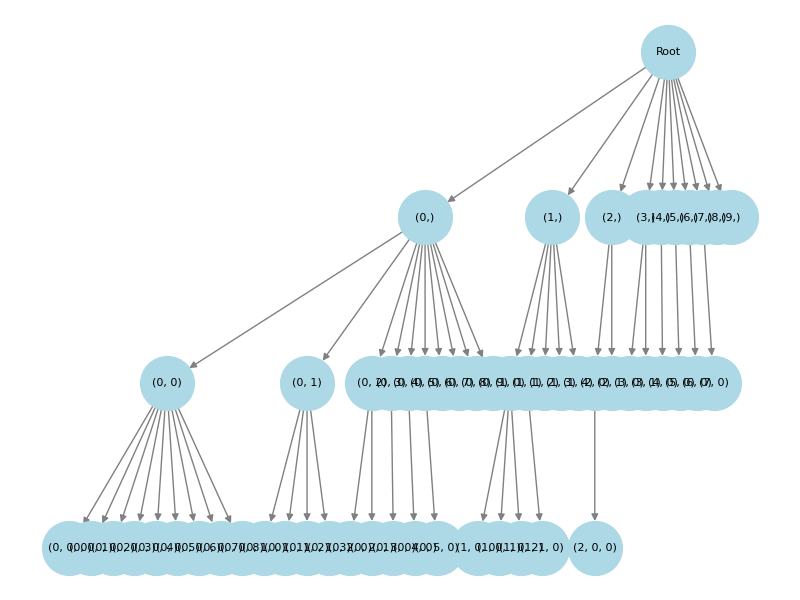} & 
    \includegraphics[width=0.18\textwidth]{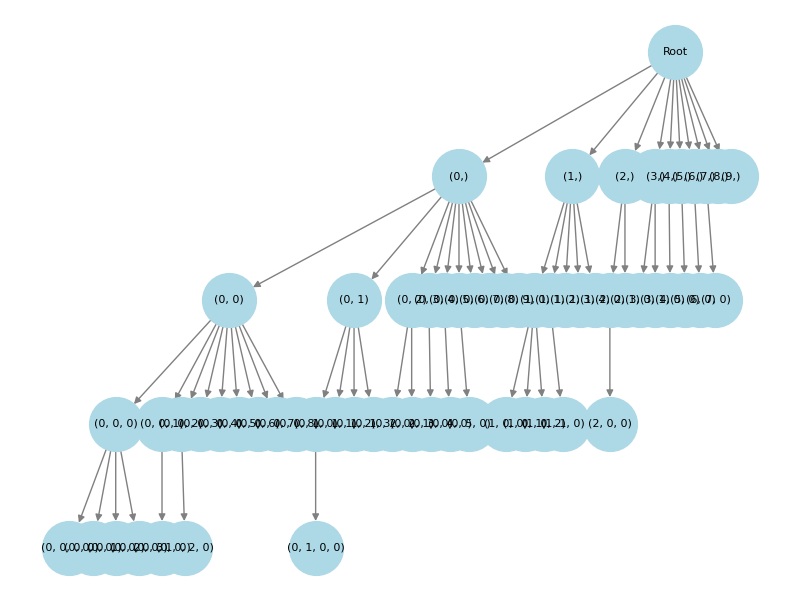} & 64 \\
    \hline
    \end{tabular}
    } 
    \label{tab:treeshapes}
\end{table}
\end{minipage}

\end{document}